\documentclass[lettersize,onecolumn]{IEEEtran}
\usepackage{amsmath,amsfonts}
\usepackage{algorithmic}
\usepackage{algorithm}
\usepackage{array}
\usepackage[caption=false,font=normalsize,labelfont=sf,textfont=sf]{subfig}
\usepackage{textcomp}
\usepackage{stfloats}
\usepackage{url}
\usepackage{verbatim}
\usepackage{graphicx}
\usepackage{cite}
\usepackage{hyperref}

\usepackage{mathrsfs} % \mathscr
\usepackage{amssymb}
\usepackage{mathtools}
\usepackage{amsthm}
\usepackage{multirow}
\usepackage{arydshln}
\usepackage{makecell}
\usepackage{bm}

\theoremstyle{plain}
\newtheorem{theorem}{Theorem}[section]

\newtheorem{lemma}[theorem]{Lemma}

\theoremstyle{definition}
\newtheorem{definition}[theorem]{Definition}

\theoremstyle{remark}

\newtheorem{restatetheoreminner}{\bf Theorem}
\newenvironment{restatetheorem}[1]{%
  \renewcommand\therestatetheoreminner{#1}%
  \restatetheoreminner
}{\endrestatetheoreminner}

\begin{document}
\title{Information-Theoretic Generalization Bounds of Replay-based Continual Learning}

\author{Wen Wen, Tieliang Gong, Yunjiao Zhang, Zeyu Gao, Weizhan Zhang, ~\IEEEmembership{Senior Member,~IEEE}, and Yong-Jin Liu, ~\IEEEmembership{Senior Member,~IEEE}
        % <-this % stops a space
\thanks{This work was supported by the Fundamental Research Funds for the Central Universities No. xxj032025002 and the Key Research and Development Project in Shaanxi Province No. 2024PT-ZCK-89. (Corresponding author: Tieliang Gong.)}% <-this % stops a space
\thanks{Wen Wen, Tieliang Gong, and Weizhan Zhang are with the School of Computer Science and Technology, Xi'an Jiaotong University, China (e-mail: wen190329@gmail.com; adidasgtl@gmail.com; zhangwzh@xjtu.edu.cn)}
\thanks{Zeyu Gao is with Department of Oncology, University of Cambridge, U.K. (e-mail: betpotti@gmail.com)}
\thanks{Yunjiao Zhang is with China Telecom, China (e-mail: zhangyj33@chinatelecom.cn)}
\thanks{Yong-Jin Liu is with  Department of Computer Science and Technology, Tsinghua University, China (e-mail: liuyongjin@tsinghua.edu.cn)}
}

\markboth{IEEE Transactions}%
{Title}

% \IEEEpubid{0000--0000~\copyright~2023 IEEE}

\maketitle

\begin{abstract}
        Continual learning (CL) has emerged as a dominant paradigm for acquiring knowledge from sequential tasks while avoiding catastrophic forgetting. Although many CL methods have been proposed to show impressive empirical performance, the theoretical understanding of their generalization behavior remains limited, particularly for replay-based approaches. This paper establishes a unified theoretical framework for replay-based CL, deriving a series of information-theoretic generalization bounds that explicitly elucidate the impact of the memory buffer alongside the current task on generalization performance. Specifically, our hypothesis-based bounds capture the trade-off between the number of selected exemplars and the information dependency between the hypothesis and the memory buffer. Our prediction-based bounds yield tighter and computationally tractable upper bounds on the generalization error by leveraging low-dimensional variables. Theoretical analysis is general and broadly applicable to a wide range of learning algorithms, exemplified by stochastic gradient Langevin dynamics (SGLD) as a representative method.  Comprehensive experimental evaluations demonstrate the effectiveness of our derived bounds in capturing the generalization dynamics in replay-based CL settings.
\end{abstract}


\section{Introduction}
Continual learning (CL) aims to incrementally learn a sequence of non-stationary tasks while retaining previously obtained knowledge \cite{parisi2019continual,wang2024comprehensive}. However, the continual acquisition of new knowledge tends to overwrite or modify learned information to accommodate evolving data, resulting in catastrophic forgetting whereby the model's performance on earlier tasks significantly deteriorates \cite{mccloskey1989catastrophic,goodfellow2013empirical}. In recent years, extensive efforts have been made to combat catastrophic forgetting in CL \cite{kao2021natural,liang2023adaptive,lyu2023overcoming}, with replay-based methods emerging as a predominant strategy that mitigates forgetting by leveraging a memory buffer of exemplars from previous tasks for experience replay \cite{guo2022online,channappayya2023augmented}. Despite its empirical success, there is a lack of theoretical understanding of the generalization characteristics of replay-based CL.

The generalization analysis of replay-based CL faces two-fold fundamental challenges, primarily attributed to stringent memory constraints and the non-stationary distributions across sequential tasks. Prior work \cite{riemerlearning,shi2023unified} has utilized replay loss with trade-off parameters to approximate the overall risk over all tasks and establish generalization bounds under the domain adaptation framework. However, they fail to explicitly disentangle the intrinsic generalization error evaluated on the full data from the approximation cost induced by memory compression, thereby yielding an insufficient characterization of the information loss inherent to finite memory capacities and overlooking the impact of memory constraints on model performance. Furthermore, the aforementioned work hinges on the complexity of hypothesis spaces, resulting in computationally intractable and vacuous bounds when applied to deep neural networks. While several recent studies have explored the theoretical underpinnings of CL by analyzing the convergence properties of learning algorithms \cite{knoblauch2020optimal,han2023convergence,li2024stability}, these results rely on stringent assumptions such as Lipschitz continuity, smoothness, and convexity, and thus provide limited insights into the generalization dynamics of replay-based CL.

To address these fundamental challenges, we employ a novel error decomposition framework that separates the generalization error into the ideal risk over the full dataset of all sequential tasks and the memory compression cost stemming from approximating the risk of the full data stream with limited memorized samples. On this basis, we then establish generalization error bounds by explicitly quantifying the dependency between the output hypothesis and the sequence of tasks, which simultaneously accounts for the data distribution and learning algorithms, while operating under substantially milder assumptions. In summary, our contributions are stated as follows:

\begin{itemize}
	\item We provide the first information-theoretic generalization bounds for replay-based CL. The resulting bounds, characterized by the accumulation of mutual information (MI) between model parameters and the memory buffer, as well as the current task data, exhibit a standard scaling rate of $\max\{\mathcal{O}(\frac{1}{\sqrt{\tilde{n}}}),\mathcal{O}(\frac{1}{\sqrt{n}})\}$ with respect to the memory size $\tilde{n}$ and the current task data size $n$. Our results underscore the necessity of retaining a sufficient number of selected exemplars while simultaneously maintaining low information dependency between parameters and memorized samples to achieve good generalization.
	
	\item We develop a family of refined bounds by quantifying the information associated with learned parameters, loss pairs, and loss differences under the supersample setting. The derived bounds benefit from the lower dimensionality of the introduced random variables, yielding computationally tractable and more stringent generalization guarantees, particularly for deep learning scenarios. Additionally, we achieve fast convergence rates of $\max\{\mathcal{O}(\frac{1}{\tilde{n}}),\mathcal{O}(\frac{1}{n})\}$ via weighted generalization error analysis, which is valid regardless of whether the empirical risk approaches zero.
	
	\item Our analysis is general and applicable to a wide range of learning algorithms. In particular, we derive novel data-dependent bounds for iterative and noisy learning algorithms in replay-based CL scenarios. The resulting bounds depend on the expected conditional gradient variance instead of conventional Lipschitz constants, enabling a more precise characterization of the optimization trajectory.
	
	\item We empirically validate our theoretical results through comprehensive experiments on the real-world datasets, demonstrating a close alignment between the derived bounds and the true generalization error across various deep learning configurations. 
\end{itemize}

\section{Related Work}
\subsection{Replay-based Continual Learning}
Replay-based methods, which jointly leverage stored data from previous tasks and newly encountered samples, have emerged as a dominant paradigm for mitigating catastrophic forgetting \cite{rebuffi2017icarl,borsos2020coresets,zhang2022simple}. Given the stringent memory constraints, the principled construction and effective utilization of the memory buffer are crucial for maximizing the efficacy of replay-based methods. A commonly adopted strategy for exemplar selection is reservoir sampling \cite{buzzega2020dark,brignac2023improving}, which ensures that each previously encountered example has an equal probability of being retained in the memory buffer during training. Alternative approaches include class-prototype sampling in feature space \cite{fan2021flar,wang2023few}, gradient-based selection \cite{aljundi2019gradient,borsos2020coresets}, and generative replay using auxiliary models \cite{shin2017continual,gao2023ddgr,liu2024tactclnet}. Despite the impressive empirical performance, a comprehensive theoretical understanding of the generalization properties of replay-based CL remains to be explored.

Several recent studies have attempted to provide preliminary theoretical insights into memory mechanisms in CL \cite{knoblauch2020optimal,han2023convergence,li2024stability,pentina2015multi,shi2023unified}. Knoblauch et al. \cite{knoblauch2020optimal} indicate that optimal CL algorithms that prevent catastrophic forgetting generally correspond to solving NP-hard problems and necessitate perfect memory. Evron et al. \cite{evron2022catastrophic} provide a theoretical characterization of the worst-case catastrophic forgetting in overparameterized linear regression models. Shi and Wang \cite{shi2023unified} develop a unified generalization bound by leveraging VC dimension, allowing different CL methods to be equipped with the same error bound up to different constants as fixed coefficients. Ding et al. \cite{ding2024understanding} provide theoretical insights into the generalization and forgetting of multi-step SGD algorithms, indicating that both the sample size and step size play critical roles in mitigating forgetting. However, none of the aforementioned work provide rigorous generalization guarantees for replay-based CL, particularly with respect to dependencies in both data distribution and learning algorithms.

\subsection{Information-theoretical Analysis}
Information-theoretic metrics are first introduced in \cite{xu2017information,russo2019much} to characterize the generalization error of learning algorithms via the mutual information between the output parameter and the input data. This methodology has been utilized in recent work to analyze the generalization dynamics of iterative and noisy learning algorithms, as exemplified by its application to SGLD \cite{negrea2019information,wang2021analyzing} and SGD \cite{neu2021information,wang2021generalization}. Subsequent significant advances involved establishing more rigorous information-theoretic generalization bounds through a variety of analytical techniques, including conditioning \cite{hafez2020conditioning}, the random subsets \cite{bu2020tightening,rodriguez2021random}, and conditional information measures \cite{steinke2020reasoning,haghifam2020sharpened}. Recent work \cite{hellstrom2022new,wang2023tighter} has developed the refined bounds by leveraging the loss pairs and loss differences, thereby yielding computational tractability and tighter generalization guarantees. However, these analyses are primarily confined to single-task learning scenarios, and their extension to continual learning settings remains unexplored.

% data-dependent or algorithm dependent 

\section{Preliminaries}
\subsection{Basic Notations}
We denote random variables by capital letters (e.g., $X$), specify their realizations using lowercase letters (e.g., $x$), and identify the domain with calligraphic letters (e.g., $\mathcal{X}$). We use the superscript $X^{i:j}$ to represent the sequence of variables $\{X^i,\ldots, X^j\}$. Let $P_{X}$ denote the probability distribution of $X$, $P_{X,Y}$ be the joint distribution of two variables $(X,Y)$, $P_{X|Y}$ be the conditional distribution of $X$ given $Y$, and $P_{X|y}$ be the conditional distribution conditioning on a sepcific value $Y=y$. Further denote by $\mathbb{E}_X$, $\mathrm{Var}_X$, and $\mathrm{Cov}_X$ the expectation, variance, and covariance matrix, respectively, for $X\sim P_X$. Given measures $P$ and $Q$, the KL divergence of $P$ with respect to (w.r.t) $Q$ is defined as $D(P\Vert Q) = \int \log\frac{dP}{dQ} dP$. For two Bernoulli distributions with $p$ and $q$, we further refer to $d(p\Vert q)=p\log(\frac{p}{q}) + (1-p)\log(\frac{1-p}{1-q})$ as the binary KL divergence. Let $H(X)$ be Shannon's entropy of variable $X$, $I(X;Y)$ be the MI between variables $X$ and $Y$, and $I(X;Y|Z) = \mathbb{E}_Z[I^Z(X;Y)] $ be the conditional mutual information (CMI) conditioned on $Z$, where $I^Z(X;Y)=D(P_{X,Y|Z}\Vert P_{X|Z}P_{Y_Z})$ denotes the disintegrated MI. Let $I_d$ be a $d$-dimensional vector of all ones. 

\subsection{Replay-based Continual Learning}
Let $\mathcal{Z}=\mathcal{X}\times\mathcal{Y}$ denote the feature-label space. Consider a sequence of $T$ learning tasks $D^{1:T}=\{D^1,\ldots, D^T\}$ arriving incrementally. For each task $i\in[T]$, $D^i=\{Z^i_j\}_{j=1}^{n}\in\mathcal{Z}^n$ consists of $n$ i.i.d. samples drawn from an unknown data-generating distribution $\mu_i$. At time $t\in[T]$, the learner has access to the current $t$-th task's data $D^t$, along with a finite memory buffer $M^{1:t-1} =\{M^i\}_{i=1}^{t-1}$, where $M^i=\{Z^i_j\}_{j=1}^{\tilde{n}}$ is a small subset ($\tilde{n}\ll n$) randomly selected from $D^i$. Let $\mathcal{W}$ denote the parameter space. Given the loss function $\ell:\mathcal{W}\times\mathcal{Z}\rightarrow\mathbb{R}^+$, the empirical risk of the hypothesis $W\in\mathcal{W}$ over all available data after each task $t$ arrives is defined by
\begin{equation}
	\hat{R}_Z(W) =  \sum_{i=1}^{t-1} \frac{1}{\tilde{n}}\sum_{j=1}^{\tilde{n}} \ell(W,Z^{i}_{j}) + \frac{1}{n}\sum_{j=1}^n \ell(W,Z^{t}_{j}).
\end{equation}
The corresponding population risk is defined by
\begin{equation}
	R(W) = \sum_{i=1}^t \mathbb{E}_{Z\sim \mu_i}[\ell(W,Z)].
\end{equation}
The discrepancy between population risk and empirical risk is known as the generalization error, denoted by 
\begin{align}\label{gener_error}
	\mathrm{gen}_{W}  \triangleq &  \mathbb{E}_{W,D^{1:t}}[R(W) - \hat{R}_Z(W)] \nonumber\\
	= & \sum_{i=1}^{t-1}\mathbb{E}_{W,D^{i}}\Big[\mathbb{E}_{Z\sim \mu_i}[\ell(W,Z)] - \frac{1}{\tilde{n}}\sum_{j=1}^{\tilde{n}} \ell(W,Z^{i}_{j})\Big] + \mathbb{E}_{W,D^{t}}\Big[ \mathbb{E}_{Z\sim \mu_t}[\ell(W,Z)] - \frac{1}{n}\sum_{j=1}^n \ell(W,Z^{t}_{j}) \Big],
\end{align}
which serves as an indicator of the output hypothesis's performance under unseen data from the distribution $\{\mu_1,\ldots,\mu_t\}$. It is worth noting that for previous task $i\in[t-1]$, its individual generalization error can be further decomposed into:
\begin{align}\label{gener_decom}
	&\mathbb{E}_{W,D^{i}}\Big[\mathbb{E}_{Z\sim \mu_i}[\ell(W,Z)] - \frac{1}{\tilde{n}}\sum_{j=1}^{\tilde{n}} \ell(W,Z^{i}_{j})\Big] \nonumber\\
	= &  
	\underbrace{\mathbb{E}_{W,D^{i}}\Big[\mathbb{E}_{Z\sim \mu_i}[\ell(W,Z)] - \frac{1}{n}\sum_{j=1}^n \ell(W,Z^{i}_{j}) \Big]}_{\text{Ideal Gap}} + \underbrace{\mathbb{E}_{W,D^{i}}\Big[  \frac{1}{n}\sum_{j=1}^n \ell(W,Z^{i}_{j}) -  \frac{1}{\tilde{n}}\sum_{j=1}^{\tilde{n}} \ell(W,Z^{i}_{j})\Big]}_{\text{Memory Compression Gap}},
\end{align}
where the Ideal Gap reflects the learning behavior when the full dataset of task $i$ is available, while the Memory Gap characterizes the approximation error introduced by using a limited number of exemplars. Consequently, the generalization error (\ref{gener_error}) together with the partial decomposition in (\ref{gener_decom}), will serve as the basis for analyzing the generalization behavior of replay-based CL. To simplify notation, we denote the expected empirical and population risks by $R\triangleq \mathbb{E}_{W}[R(W)]$ and $\hat{R}\triangleq \mathbb{E}_{W,D^{1:t}}[ \hat{R}_Z(W)]$, respectively.

\subsection{Supersample Setting}
We utilize the CMI framework \cite{steinke2020reasoning} for a rigorous generalization analysis. Let $\tilde{D}^{i}=\{\tilde{Z}^i_j\}_{j=1}^n\in\mathcal{Z}^{n\times 2}$ be the supersample dataset for each task $i\in[T]$ consisting of $2n$ i.i.d. samples drawn from $\mu_i$, where each element $\tilde{Z}^i_j =(\tilde{Z}^i_{j,0},\tilde{Z}^i_{j,1})$ comprises a pair of samples. Let $S^i=(S^i_1,\ldots, S^i_n)\sim\mathrm{Unif}(\{0,1\}^{ n})$, independent of $\tilde{D}^{i}$, be used to segregate training and test samples from $\tilde{D}^{i}$, namely, $S^i_j = 0$ indicates that $\tilde{Z}^i_{j,0}$ is used for training and $\tilde{Z}^i_{j,1}$ for testing. The corresponding training and test sets are represented by $\tilde{D}^{i}_{S^i}=\{\tilde{Z}^i_{j,S^i_j}\}_{j=1}^{n}$ and $ \tilde{D}^{i}_{\bar{S}^i}=\{\tilde{Z}^i_{j,1-S^i_j}\}_{j=1}^{n}$, respectively. Further let $U^i=(U^i_1,\ldots, U^i_n)\in \{0,1\}^n$ be a binary vector satisfying $\sum_{j=1}^n U^i_j = \tilde{n}$, used to select the memory $\tilde{M}^i$ from the training dataset $\tilde{D}^{i}_{S^i}$ such that $\tilde{M}^i=\{\tilde{Z}^i_{j,S^i_j}\in \tilde{D}^{i}_{S^i} | U^i_j =1  \}$. We denote the sample-wise losses on the training and test data by $L^i_{j,S^i_j}=\ell(W,Z^i_{j,S^i_j})$ and $L^i_{j,\bar{S}^i_j}=\ell(W,Z^i_{j,1-S^i_j})$, respectively. $L^i_j=\{L^i_{j,0},L^i_{j,1}\}$ then represents a pair of losses, and $\Delta^i_j = L^i_{j,1} -L^i_{j,0}$ is their difference. The collection of sample-wise training losses is defined as $L^{i,S^i}=\{L^i_{j,S^i_j}\}_{j=1}^n$, and $L^{i,\bar{S}^i}$ defined similarly.

\section{Hypothesis-based Generalization Bounds}\label{Section3}
In this section, we provide theoretical insights into the generalization properties of replay-based continual learning by analyzing the dependence between the learned hypothesis and the available data. Our theoretical results elucidates the impact of the compressed and informative memory buffer on generalization performance. Detailed proofs for the theorems are presented in the Appendix.
\subsection{Input-output MI Bounds}
We begin by deriving a generalization bound based on input-output MI, expressed through the mutual information between the model parameters $W$ and the current task data $D^t$, as well as that between $W$ and the memory buffer $M^{1:t-1}$:
\begin{theorem}\label{input-output}
	Let $n$ and $\tilde{n}$ denote the number of samples available for training the current task $D^t$ and the number of samples from the previous task $i$ in memory $M^i$, respectively, where $t\in[T]$ and $i\in[t-1]$. Assume that $\ell(w,Z)$, where $Z\in\mathcal{Z}$, is $\sigma$-subgaussian for all $w\in\mathcal{W}$, we have 
	\begin{equation*}
		\vert \mathrm{gen}_{W} \vert \leq   \underbrace{\sqrt{\frac{2\sigma^2}{n} I(W;D^t)}}_{\text{Current Task Generalization}} + \sum_{i=1}^{t-1} \bigg(\underbrace{\sqrt{\frac{2\sigma^2}{n} I(W;M^i)}}_{\text{Previous Task Generalization}} + \underbrace{\sqrt{\frac{2\sigma^2(n-\tilde{n})}{n\tilde{n}} I(W;M^i)}}_{\text{Memory Compression Cost}}\bigg).
	\end{equation*}   
\end{theorem}
Theorem \ref{input-output} explicitly disentangles the generalization bound of replay-based CL into three governing terms: \textit{Current Task Generalization}, \textit{Previous Task Generalization}, and \textit{Memory Compression Cost}. Specifically, both the \textit{Current Task Generalization} and \textit{Previous Task Generalization} terms characterize the model's generalization performance under access to full datasets, scaling at the standard rate of $\mathcal{O}(1/\sqrt{n})$ with respect to the sample size $n$. As only limited exemplars are available for previous tasks $i\in[t-1]$, the \textit{Previous Task Generalization} term is governed by the dependency between $W$ and the memory $M^i$ rather than the entire dataset $D^i$, where the conditional independence $I(W;D^i\backslash M^i|M^i)=0$ leads to $I(W; D^i)=I(W;M^i)$. This implies that the less the learned hypothesis $W$ depends on the input data $M^{1:t-1}$ or $D^t$, the better its generalization performance, which aligns with the general consensus in \cite{xu2017information}. Additionally, the \textit{Memory Compression Cost} term, scaling as $\mathcal{O}(\sqrt{\frac{n-\tilde{n}}{n\tilde{n}}})$ (asymptotically $\mathcal{O}(1/\sqrt{\tilde{n}})$ when $n\gg \tilde{n}$), captures a fundamental trade-off between memory capacity $\tilde{n}$ and information complexity $I(W;M^i)$: minimizing this term requires increasing the exemplar size $\tilde{n}$ while simultaneously constraining the MI $I(W;M^i)$. Specifically, while a larger $\tilde{n}$ reduces the scaling factor $\mathcal{O}(1/\sqrt{\tilde{n}})$, it typically induces higher dependency $I(W;M^i)$ as the parameters $W$ tend to fit the stored exemplars $M^i$ well, thereby not necessarily leading to a decreased memory compression term. This thus suggests that it is crucial to leverage appropriate sample selection strategies to carefully choose a sufficient number of representative exemplars $M^i$ with low information dependence $I(W;M^i)$, thereby ensuring small generalization errors.

\subsection{CMI Bounds}
Building upon the CMI framework \cite{steinke2020reasoning}, we establish the following refined generalization bound that improves the MI upper bound and quantifies the impact of the constructed memory buffer on the generalization performance:
\begin{theorem}\label{cmi}
	Let $n$ and $\tilde{n}$ denote the number of samples available for training the current task and the number of the memory, respectively. Assume that $\ell(w,Z)$, where $Z\in\mathcal{Z}$, is $\sigma$-subgaussian for all $w\in\mathcal{W}$, we have 
	\begin{equation*}
		\vert \mathrm{gen}_{W} \vert \leq \underbrace{\mathbb{E}_{\tilde{D}^{t}} \sqrt{\frac{4\sigma^2 I^{\tilde{D}^t}(W; S^i)}{n}}}_{\text{Current Task Generalization}} + \underbrace{\mathbb{E}_{\tilde{D}^{1:t-1}} \sum_{i=1}^{t-1} \sqrt{\frac{4\sigma^2I^{\tilde{D}^i}(W; S^i)}{n}}}_{\text{Previous Task Generalization}} + \underbrace{\mathbb{E}_{\tilde{D}^{1:t-1},S^{1:t-1}} \sum_{i=1}^{t-1}\sqrt{\frac{2\sigma^2 (n-\tilde{n})}{n\tilde{n}} I^{\tilde{D}^{i}_{S^i}}(W; U^{i})}}_{\text{Memory Compression Cost}}.
	\end{equation*}
\end{theorem}
Notably, Theorem \ref{cmi} significantly tightens the derived bound in Theorem \ref{input-output} by leveraging the CMI terms between the hypothesis $W$ and the sample selection variables $S^i$ (or memory selection variables $U^i$), conditioned on the full dataset $\tilde{D}^{i}$ (or the training dataset $\tilde{D}^{i}_{S^i}$). Specifically, for any $i\in[t]$, the Markov chain $(\tilde{D}^{1:t},S^{1:t},U^{1:t-1})-(\tilde{D}^t_{S^t},\tilde{M}^{1:t-1})-W$ implies the conditional independence $I(W;\tilde{D}^{i},S^{i}|\tilde{D}^i_{S^i})=0$, thereby yielding the inequality $I(W;\tilde{D}^i_{S^i}) = I(W;\tilde{D}^{i},S^{i}) = I(W;S^{i}| \tilde{D}^{i}) + I(W;\tilde{D}^i) \geq I(W;S^{i}| \tilde{D}^{i})$. Analogously, it can be proven that $I(W;\tilde{M}^{i}) = I(W;U^i| \tilde{D}^i_{S^i}) + I(W;\tilde{D}^i_{S^i}) \geq I(W;U^i| \tilde{D}^i_{S^i})$. Additional insights beyond Theorem \ref{input-output} highlight that minimizing the informational dependency on specific indices $S^i$ and $U^i$ is critical to achieve better generalization. This also theoretically interprets the effectiveness of random replay \cite{riemer2018learning,wang2024comprehensive}, which inherently maintains low selection dependency.

% \begin{table*}
	%   \caption{Some Typical Commands}
	%   \label{tab:commands}
	%   \begin{tabular}{ccl}
		%     \toprule
		%     Command &A Number & Comments\\
		%     \midrule
		%     \texttt{{\char'134}author} & 100& Author \\
		%     \texttt{{\char'134}table}& 300 & For tables\\
		%     \texttt{{\char'134}table*}& 400& For wider tables\\
		%     \bottomrule
		%   \end{tabular}
	% \end{table*}

\subsection{Algorithm-based Bounds}
We further investigate the generalization properties of noisy iterative learning algorithms in the context of CL \cite{wang2023distributionally,chen2023stability}, with a focus on stochastic gradient Langevin dynamics (SGLD). At each time $t\in[T]$, we denote the training trajectory of SGLD over the current task $D^t$ and memory buffer $M^{1:t-1}$ as $\{W_r\}_{r=0}^R$, where $W_0\in\mathbb{R}^d$ represents the initialized model parameters. In the $r$-th update, we construct the mini-batches $\{\tilde{B}^r_1,\ldots, \tilde{B}^r_{t-1}, B^r_t\}$ by independently sampling $\tilde{B}^r_i \subseteq M^i$ with $\vert \tilde{B}^r_i\vert = \tilde{b}_i$ for all $i\in[t-1]$, alongside a batch $B^r_t\subseteq D^t$ of size $b_t$ from the current task. The update rule of model parameters can then be formalized by
\begin{equation*}
	W_r = W_{r-1} - \eta_r \left(\sum_{i=1}^{t-1}\frac{1}{\tilde{b}_i} \sum_{Z\in \tilde{B}^r_i} \nabla \ell(W_r,Z)  + \frac{1}{b_t} \sum_{Z\in B^r_t} \nabla \ell(W_r,Z)  \right) +N_r,
\end{equation*}
where  $\eta_r$ denotes the learning rate and $N_r\sim N(0,\xi^2_r I_d)$ is the isotropic Gaussian noise injected in each step. To simplify the notation, we denote $\tilde{g}_i(W_r) = -\frac{1}{\tilde{b}_i} \sum_{Z\in \tilde{B}^r_i} \nabla \ell(W_r,Z)$ for all $i\in[t-1]$ and $g_t(W_r) = -\frac{1}{b_t} \sum_{Z\in B^r_t} \nabla \ell(W_r,Z)$.

In the following theorem, we provide data-dependent bounds for SGLD-based CL algorithms by leveraging the determinant trajectory of the gradient covariance matrices with respect to the memory buffer $M^{1:t-1}$ and the current task data $D^t$:
\begin{theorem}\label{algorithm}
	Let $n$ and $\tilde{n}$ denote the number of samples available for training the current task and the number of the memory, respectively. Assume that $\ell(w,Z)$, where $Z\in\mathcal{Z}$, is $\sigma$-subgaussian for all $w\in\mathcal{W}$. Let $W$ be the output of the SGLD algorithm after $R$ iterations at time $t$, then
	\begin{equation*}
		\vert \mathrm{gen}_{W} \vert \leq  \sum_{i=1}^{t-1} \sqrt{\frac{2\sigma^2}{\tilde{n}} \sum_{r=1}^{R}\log \Big\vert \frac{\eta_r^2}{\xi^2_r} \mathbb{E}_{W_{r-1}}\big[ \tilde{\Sigma}^r_i \big]  + I_d \Big\vert} +  \sqrt{\frac{\sigma^2}{n} \sum_{r=1}^{R} \log \Big\vert \frac{\eta_r^2}{\xi^2_r} \mathbb{E}_{W_{r-1}}\big[ \Sigma^r_t \big]  + I_d \Big\vert} ,
	\end{equation*} 
	where $\tilde{\Sigma}^r_i = \mathrm{Cov}_{\tilde{B}^r_i}[\tilde{g}_i(W_r)]$ for all $i\in[t-1]$, and $\Sigma^r_t = \mathrm{Cov}_{B^r_t}[g_t(W_r)]$.
\end{theorem}
Theorem \ref{algorithm} precisely characterizes the generalization dynamics of SGLD-based CL algorithms through the accumulated gradient covariance $\{\tilde{\Sigma}^r_1.\ldots,\tilde{\Sigma}^r_{t-1},\Sigma^r_t\}$ and predefined quantities such as learning rates $\eta_r$, variance $\xi_r$, and number $R$ of iterations. Compared with previous work \cite{chen2023stability,bonicelli2022effectiveness}, the derived bound leverages the gradient covariance matrix rather than the Lipschitz constant of the loss function, thereby yielding tighter and non-vacuous generalization guarantees, particularly in deep learning scenarios. Our result also indicates that the average gradient over both the memory buffer and the current task data should be as small as possible to achieve smaller generalization errors, which facilitates the learning parameters to converge toward the stationary points of the empirical risk landscape \cite{gupta2020look,deng2021flattening}.

\section{Prediction-based Generalization Bounds}\label{Section4}
In this section, we establish tighter generalization bounds for replay-based continual learning by shifting the analysis from parameter-level dependencies to the information captured by the loss function. This analysis further elucidates the interaction between the generalization error and the empirical losses evaluated over the memory buffer and the current task data.
\subsection{Loss-based Bounds}
In the following theorem, we present the loss-based information-theoretic bound that incorporates the CMI between the evaluated losses and the selection variables conditioned on the supersamples, yielding a provably tighter upper bound:

\begin{theorem}\label{e-cmi}
	Let $n$ and $\tilde{n}$ denote the number of samples available for training the current task and the number of the memory, respectively. Assume that $\ell(w,Z)$, where $Z\in\mathcal{Z}$, is $\sigma$-subgaussian for all $w\in\mathcal{W}$, we have 
	\begin{equation*}
		\vert \mathrm{gen}_{W} \vert \leq \underbrace{\mathbb{E}_{\tilde{D}^{t}} \sqrt{\frac{4\sigma^2 I^{\tilde{D}^t}(L^{i}; S^i)}{n}}}_{\text{Current Task Generalization}} + \underbrace{\mathbb{E}_{\tilde{D}^{1:t-1}} \sum_{i=1}^{t-1} \sqrt{\frac{4\sigma^2I^{\tilde{D}^i}(L^{i}; S^i)}{n}}}_{\text{Previous Task Generalization}} + \underbrace{\mathbb{E}_{\tilde{D}^{1:t-1},S^{1:t-1}} \sum_{i=1}^{t-1}\sqrt{\frac{2\sigma^2 (n-\tilde{n})}{n\tilde{n}} I^{\tilde{D}^{i}_{S^i}}(L^{i}; U^{i})}}_{\text{Memory Compression Cost}}.
	\end{equation*}
\end{theorem}
Theorem \ref{e-cmi} links the generalization error to the CMI involving the loss pairs $L^i=\{L^i_{j,0},L^i_{j,1}\}_{j=1}^n$, leading to a tighter and more computationally tractable bound than the hypothesis-based counterpart established in Theorem \ref{cmi}. By applying the Data Processing Inequality (DPI) to the Markov chain $(S^i, U^i) - W - L^i$ conditioned on $\tilde{D}^{i}$, it is evident that the mutual information $I(L^i;S^i|\tilde{D}^{i})$ and $I(L^{i}; U^{i}|\tilde{D}^{i}_{S^i})$ serve as a tighter bound on the objectives $I(W;S^i|\tilde{D}^{i})$ and $I(W; U^{i}|\tilde{D}^{i}_{S^i})$, thereby improving upon the existing results. Furthermore, this result corroborates the benefit of reducing the dependency between loss evaluations and the input samples to improve generalization, aligning with the findings in \cite{deng2021flattening}.

Analogously to Theorem \ref{e-cmi}, we derive an unconditional upper bound governed by the mutual information between the loss and the selection variables, without the conditioning on supersamples:
\begin{theorem}\label{e-mi}
	Let $n$ and $\tilde{n}$ denote the number of samples available for training the current task and the number of the memory, respectively. Assume that $\ell(w,Z)$, where $Z\in\mathcal{Z}$, is $\sigma$-subgaussian for all $w\in\mathcal{W}$, we have 
	\begin{equation*}
		\vert \mathrm{gen}_{W} \vert \leq \underbrace{\sqrt{\frac{4\sigma^2 I(L^i; S^{i})}{n}}}_{\text{Current Task Generalization}} + \sum_{i=1}^{t-1} \Bigg( \underbrace{ \sqrt{\frac{4\sigma^2I(L^i; S^i)}{n}}}_{\text{Previous Task Generalization}} + \underbrace{ \sqrt{\frac{2\sigma^2 (n-\tilde{n})I(L^{i,S^i}; U^{i})}{n\tilde{n}} }}_{\text{Memory Compression Cost}} \Bigg).
	\end{equation*}
\end{theorem}
Notably, this unconditional upper bound is consistently tighter than its conditional counterpart in Theorem \ref{e-cmi}. Given the independence between $S^i$ and $\tilde{D}^{i}$ (or $U^i$ and $\tilde{D}^{i}_{S^i}$), we observe that $I(L^i;S^i)\leq I(L^i;S^i) + I(S^i;\tilde{D}^{i}|L^i) = I(L^i;S^i|\tilde{D}^{i})$ and $I(L^{i,S^i}; U^{i})\leq I(L^i,S^i; U^{i}) + I(U^i;\tilde{D}^{i}_{S^i}|L^i, S^i) = I(L^i,S^i;U^i|\tilde{D}^{i}_{S^i}) = I(L^i;U^i|\tilde{D}^{i}_{S^i})$, which demonstrates the tightness of Theorem~\ref{e-mi}. Additionally, the subsequent theorem establishes the binary KL divergence bound between the expected empirical risk $\hat{R}$ and the mean of the expected empirical and population risks $(\hat{R}+R)/2$:
\begin{theorem}\label{binarykl2}
	Let $n$ and $\tilde{n}$ denote the number of samples available for training the current task and the number of the memory, respectively. Assume that $\ell(\cdot,\cdot) \in [0,1]$, we have
	\begin{equation*}
		d\bigg(\hat{R} \bigg\Vert \frac{\hat{R}+R}{2} \bigg)  \leq \sum_{i=1}^{t-1}  \frac{1}{\tilde{n}}\sum_{j:U^i_j=1}I(L^i_j;S^i_j|U^i_j=1)  +  \frac{1}{n}\sum_{j=1}^{n} I(L^t_j;S^t_j).
	\end{equation*}
\end{theorem}
Theorem \ref{binarykl2} exhibits a favorable fast scaling rate of $\mathcal{O}(\max\{\frac{1}{n},\frac{1}{\tilde{n}}\})$ with respect to the memory size $\tilde{n}$ and current task sample size $n$. Furthermore, this bound aligns with the principle of pointwise stability, which posits that algorithms stable against individual data points yield better generalization performance. While the derived bounds in Theorems \ref{e-mi} and \ref{binarykl2} achieve substantial improvements over hypothesis-based counterparts, they encompass losses evaluated across all combinations of training and test samples from the memory buffer and the current task, potentially introducing additional computational challenges in MI estimation. An alternative approach presented below is to investigate the dependence between only two one-dimensional variables (i.e., the difference of loss pairs and the selection variable), thereby establishing tighter generalization bounds.

\subsection{Loss-difference Bounds}
In the following theorem, we establish a tighter and computationally tractable bound by leveraging the loss difference $\Delta^i_j$ between the training and test samples:
\begin{theorem}\label{ld-mi}
	Let $n$ and $\tilde{n}$ denote the number of samples available for training the current task and the number of the memory, respectively. Assume that $\ell(\cdot,\cdot) \in [0,1]$, we have
	\begin{equation*}
		\vert \mathrm{gen}_W\vert \leq  \mathbb{E}_{U^{1:t-1}} \sum_{i=1}^{t-1} \frac{1}{\tilde{n}}\sum_{j:U^i_j=1} \sqrt{2I^{U^i_j=1}(\Delta^i_j;S^i_j)} + \frac{1}{n}\sum_{j=1}^{n} \sqrt{2I(\Delta^t_j;S^t_j)}.
	\end{equation*}
\end{theorem}
The quantities $I^{U^i_j=1}(\Delta^i_j;S^i_j)$ and $I(\Delta^t_j;S^t_j)$ characterize the dependency between the scalar loss difference and the selection variable for the memory data and the current task data, enabling efficient low-dimensional MI estimation. Analogous to the analysis in Theorems \ref{e-cmi} and \ref{e-mi}, it is evident that these terms provide a strictly tighter bound than prior loss-based counterparts.  For the special case of the loss function $\ell(\cdot,\cdot)\in\{0,1\}$, these terms can be interpreted as the rate of reliable communication over a memoryless channel with input $S^i_j$ (or $S^t_j$) and output $\Delta^i_j$ (or $\Delta^t_j$), thereby leading to the following fast-rate bounds in the interpolating setting: 
\begin{theorem}\label{relia-fast}
	Let $n$ and $\tilde{n}$ denote the number of samples available for training the current task and the number of the memory, respectively. Assume that $\ell(\cdot,\cdot) \in \{0,1\}$. In the interpolating setting when $\hat{R}=0$, we have 
	\begin{align*}
		R =  \sum_{i=1}^{t-1} \frac{1}{\tilde{n}}\sum_{j:U^i_j=1} \frac{I(\Delta^i_j;S^i_j|U^i_j=1)}{\log 2} + \frac{1}{n}\sum_{j=1}^{n} \frac{I(\Delta^t_j;S^t_j)}{\log 2}
		=  \sum_{i=1}^{t-1} \frac{1}{\tilde{n}}\sum_{j:U^i_j=1} \frac{I(L^i_j;S^i_j|U^i_j=1)}{\log 2} + \frac{1}{n}\sum_{j=1}^{n} \frac{I(L^t_j;S^t_j)}{\log 2}.  
	\end{align*}
\end{theorem}
Theorem \ref{relia-fast} establishes a sharp generalization bound achieving a fast rate of $\mathcal{O}(\max\{\frac{1}{n},\frac{1}{\tilde{n}}\})$, in the sense that this bound benefits from a small empirical error. In this interpolating regime, the expected population risk is precisely characterized by the summation of sample-wise MI between the selection variables and either the loss pairs or the resulting loss differences  across both memory buffer and the current task. Further refinement of Theorem \ref{relia-fast} is achievable by developing a fast-rate bound via the weighted generalization error $R - (1+C_1) \hat{R}$ (where $C_1$ is a predefined constant), which accommodates a wide range of bounded loss functions: 
\begin{theorem}\label{fast-loss}
	Let $n$ and $\tilde{n}$ denote the number of samples available for training the current task and the number of the memory, respectively. Assume that $\ell(\cdot,\cdot) \in [0,1]$, for any $0 < C_2 < \frac{\log 2}{2}$ and $C_1\geq -\frac{\log(2-e^{C_2})}{2C_2}-1$, we have
	\begin{equation*}
		\mathrm{gen}_W \leq C_1\hat{R}+ \sum_{i=1}^{t-1} \sum_{j:U^i_j=1} \frac{ \min \{I(L^i_{j};S^i_j|U^i_j=1), 2 I(L^i_{j,1};S^i_j|U^i_j=1)\} }{\tilde{n}C_2} + \sum_{j=1}^{n}\frac{\min \{I(L^t_{j};S^t_j), 2I(L^t_{j,1};S^t_j) \}}{nC_2}.
	\end{equation*}
	In the interpolating regime that $\hat{R} = 0$, we further have 
	\begin{equation*}
		R \leq  \sum_{i=1}^{t-1} \sum_{j:U^i_j=1} \frac{2 \min \{I(L^i_{j};S^i_j|U^i_j=1), 2 I(L^i_{j,1};S^i_j|U^i_j=1)\} }{\tilde{n}\log 2} + \sum_{j=1}^{n}\frac{2 \min \{I(L^t_{j};S^t_j), 2I(L^t_{j,1};S^t_j) \}}{n\log 2}.
	\end{equation*}
\end{theorem}
Theorem \ref{fast-loss} attains a convergence rate of  $\mathcal{O}(\max\{\frac{1}{n},\frac{1}{\tilde{n}}\})$ comparable to that of Theorem \ref{relia-fast}, while further tightening the bound by simultaneously accounting for the minimum between the paired-loss MI $I(L^i_{j};S^i_j |U^i_j=1 )$ (or $I(L^t_{j};S^t_j)$) and the single-loss MI $2I(L^i_{j,1};S^i_j|U^i_j=1)$ (or $2I(L^t_{j,1};S^t_j)$). The absence of a definitive ordering between the MI terms $I(L^i_{j};S^i_j|U^i_j=1)$ and $2I(L^i_{j,1};S^i_j|U^i_j=1)$ enables more flexible generalization bounds than those in Theorem \ref{relia-fast}, as the interaction information $I(L^i_{j,1};L^i_{j,0};S^i_j |U^i_j=1) = I(L^i_{j,1};S^i_j |U^i_j=1)- I(L^i_{j,1};S^i_j|L^i_{j,0},U^i_j=1) = 2I(L^i_{j,1};S^i_j|U^i_j=1) - I(L^i_{j};S^i_j|U^i_j=1) $ can take either positive or negative values. A similar analysis applies to the corresponding terms for the current task. Notably, a rigorous analysis of empirical loss variance can yield tighter fast-rate generalization bounds under the non-interpolating setting, where the loss variance, $\mathrm{Var}(\gamma)$, is defined as 
\begin{equation*}
	\mathrm{Var}(\gamma) := \mathbb{E}_{W,D^{1:T}} \bigg[ \sum_{i=1}^{t-1}\sum_{j=1}^{\tilde{n}}  \frac{\Big(\ell(W,Z^i_j)-(1+\gamma)\hat{R}^i(W)\Big)^2}{\tilde{n}} + \sum_{j=1}^{n}\frac{\Big(\ell(W,Z^t_j)-(1+\gamma)\tilde{R}^t(W)\Big)^2}{\tilde{n}} \bigg],
\end{equation*}
where $\hat{R}^i(W) = \frac{1}{\tilde{n}}\sum_{j=1}^{\tilde{n}}\ell(W,Z^i_j)$ for $i\in[t-1]$, and  $\tilde{R}^t(W)=\frac{1}{n}\sum_{j=1}^{n}\ell(W,Z^t_j)$.
\begin{theorem}\label{fast-var}
	Let $n$ and $\tilde{n}$ denote the number of samples available for training the current task and the number of the memory, respectively. Assume that $\ell(\cdot,\cdot) \in \{0,1\}$, for any $0 < C_2 < \frac{\log 2}{2}$ and $C_1\geq -\frac{\log(2-e^{C_2})}{2C_2}-1$, we have
	\begin{equation*}
		\mathrm{gen}_W  \leq C_1 \mathrm{Var}(\gamma)  + \sum_{i=1}^{t-1} \sum_{j:U^i_j=1} \frac{ \min \{I(L^i_{j};S^i_j|U^i_j=1), 2 I(L^i_{j,1};S^i_j|U^i_j=1)\} }{\tilde{n}C_2} + \sum_{j=1}^{n}\frac{\min \{I(L^t_{j};S^t_j), 2I(L^t_{j,1};S^t_j) \}}{nC_2}.
	\end{equation*}
\end{theorem}
For binary loss functions, it has been proven that $\mathrm{Var}(\gamma) = \hat{R} -(1-\gamma^2) \mathbb{E}_{W,D^{1:T}} \big[\sum_{i=1}^{t-1} \big(\hat{R}^i(W)\big)^2 + \big(\tilde{R}^t(W)\big)^2 \big]$ for any $\gamma\in(0,1)$. Replacing $\hat{R}$ with $\mathrm{Var}(\gamma)$ indicates that the resulting loss variance bound is tighter than the interpolating bound established in Theorem \ref{fast-loss} with the margin of at least $C_1(1-\gamma^2)\mathbb{E}_{W,D^{1:T}} \big[\sum_{i=1}^{t-1} \big(\hat{R}^i(W)\big)^2 + \big(\tilde{R}^t(W)\big)^2 \big]$ using the same constants $C_1$ and $C_2$. Hence, Theorem \ref{fast-var} ensures a fast convergence rate even when the empirical risk is near but not exactly zero.

% \begin{figure}[h]
	%   \centering
	%   \includegraphics[width=\linewidth]{sample-franklin}
	%   \caption{1907 Franklin Model D roadster. Photograph by Harris \&
		%     Ewing, Inc. [Public domain], via Wikimedia
		%     Commons. (\url{https://goo.gl/VLCRBB}).}
	%   \Description{A woman and a girl in white dresses sit in an open car.}
	% \end{figure}

\section{Experiments}
\begin{figure*}[t]
	\centering
	\subfloat[MNIST ($m=250$)]{\includegraphics[width=55mm]{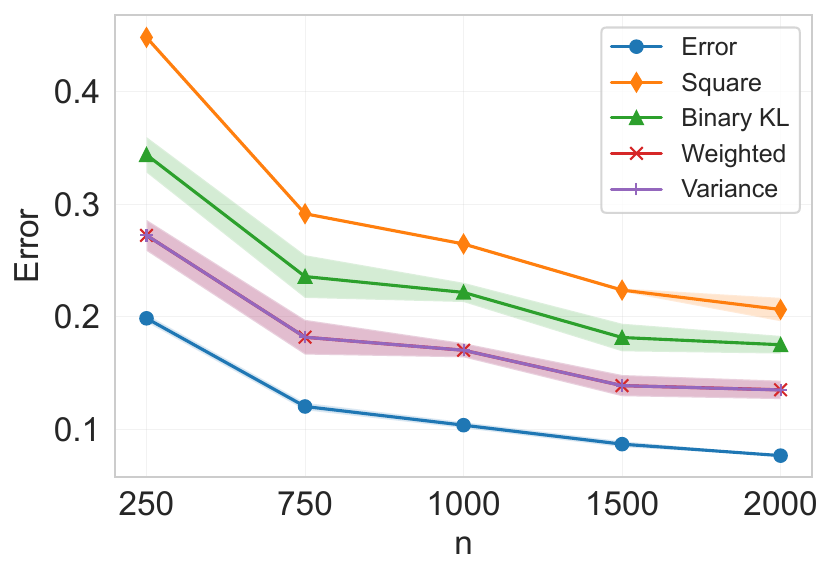}%
		\label{fig_second_case}}
	\subfloat[CIFAR-10 ($m=250$)]{\includegraphics[width=55mm]{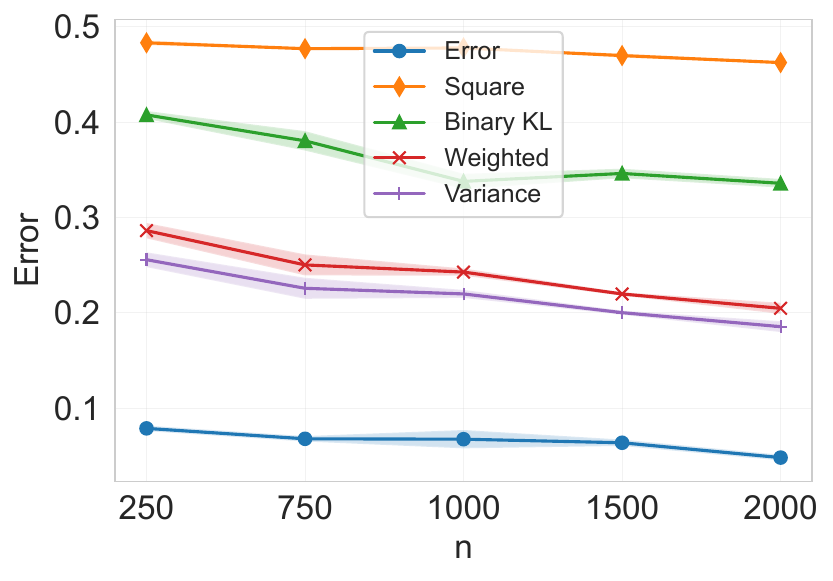}%
		\label{fig_second_case}}
	\hfil
	\subfloat[MNIST ($n=750$)]{\includegraphics[width=55mm]{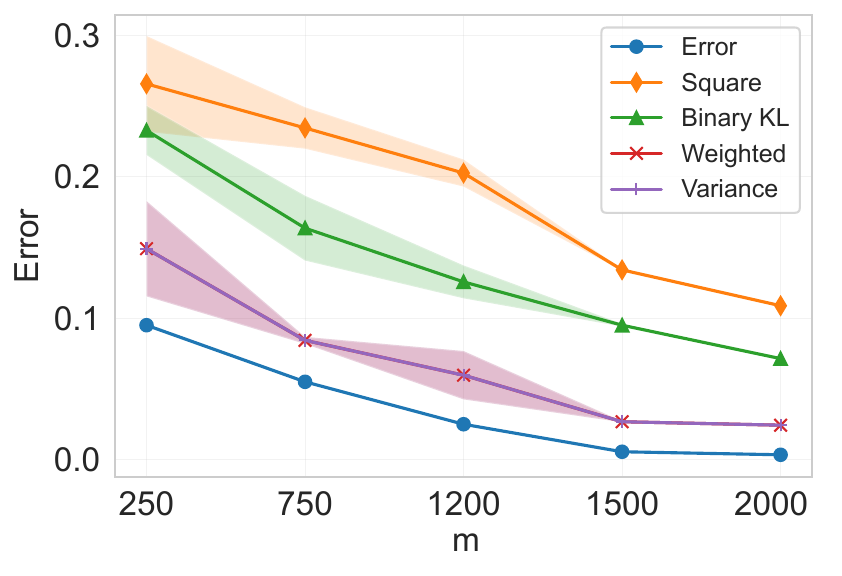}%
		\label{fig_first_case}}
	\subfloat[CIFAR-10 ($n=750$)]{\includegraphics[width=55mm]{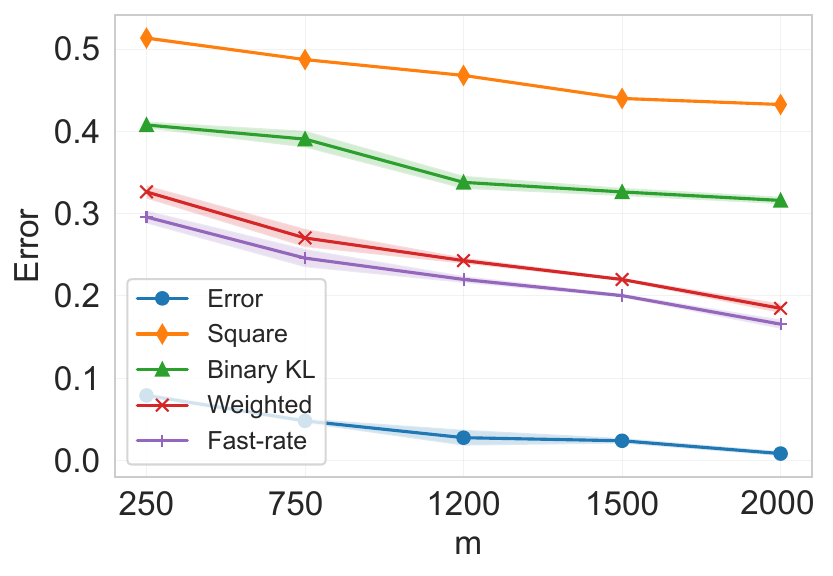}%
		\label{fig_second_case}}
	\caption{Comparison of the generalization bounds on real-world datasets under different memory buffer sizes $m$ and the number $n$ of the current task data.}
	\label{fig_1}
\end{figure*}

\begin{figure*}[t]
	\centering
	\subfloat[MNIST, CNN (SGD)]{\includegraphics[width=80mm]{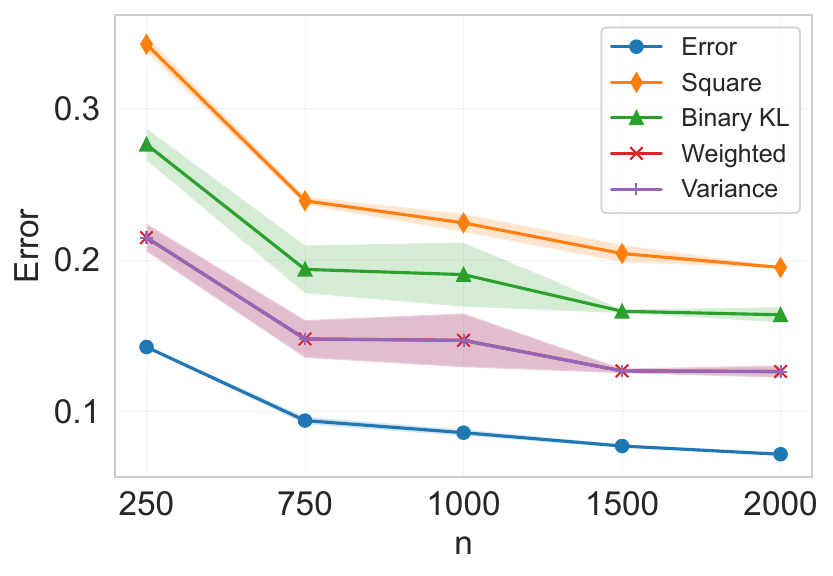}%
		\label{fig_first_case}}
	\subfloat[MNIST, CNN (SGLD)]{\includegraphics[width=80mm]{image/mnist-bu1.pdf}%
		\label{fig_second_case}}
	\hfil
	\subfloat[CIFAR-10, ResNET (SGD)]{\includegraphics[width=80mm]{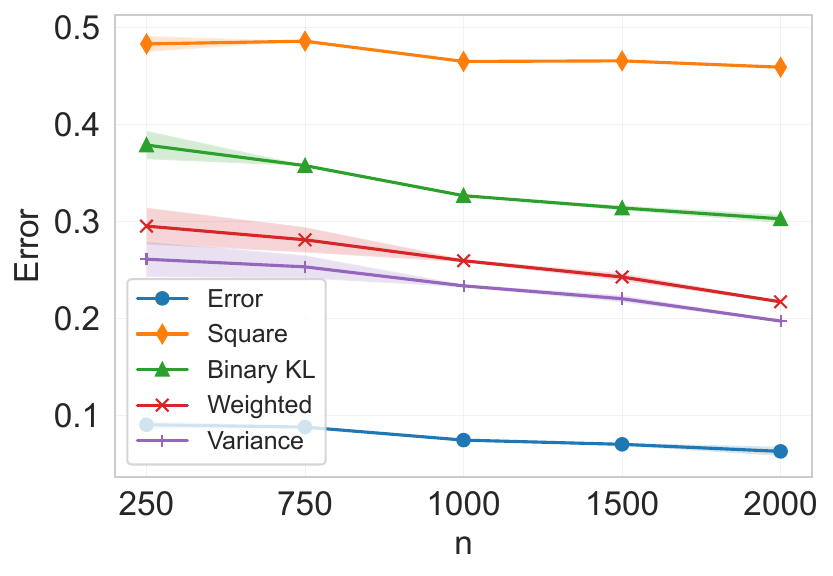}%
		\label{fig_second_case}}
	\subfloat[CIFAR-10, ResNET (SGLD)]{\includegraphics[width=80mm]{image/cifar-bu2.pdf}%
		\label{fig_second_case}}
	\caption{Comparison of the generalization bounds in multiple real-world learning scenarios under fixed memory buffer size $m=400$.}
	\label{fig_2}
\end{figure*}

\begin{figure*}[t]
	\centering
	\subfloat[$\eta=0.05$, $\theta=6$]{\includegraphics[width=60mm]{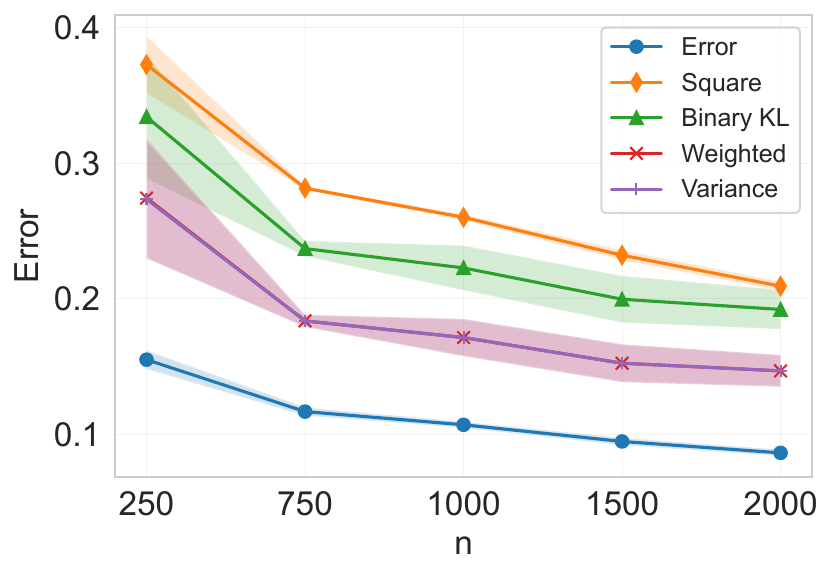}%
	}
	\subfloat[$\eta=0.01$, $\theta=8$]{\includegraphics[width=60mm]{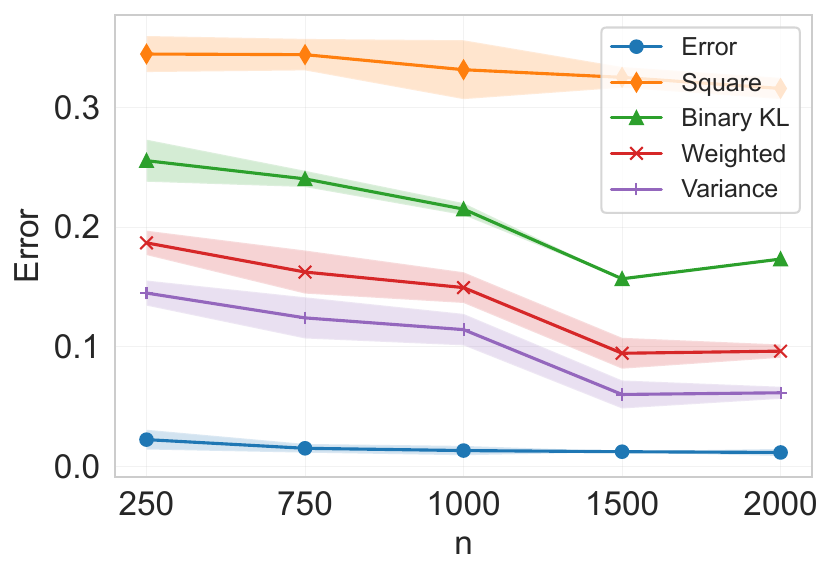}%mnist_01_8.pdf
	}
	\subfloat[$\eta=0.005$, $\theta=9$]{\includegraphics[width=60mm]{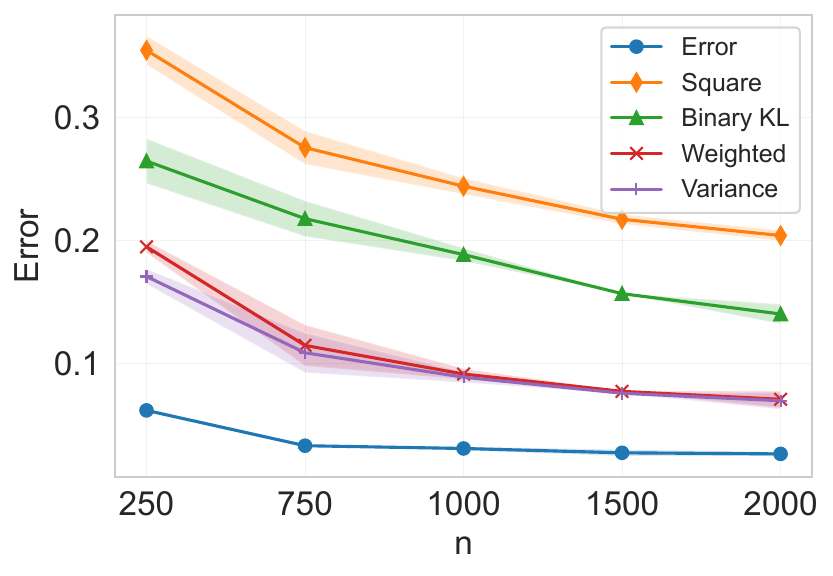}%
	}
	\caption{Comparison of the generalization bounds for the SGLD algorithm on the MNIST dataset with different learning rates $\eta$ and noise variances $\theta$.}
	\label{fig_3}
\end{figure*}

\begin{figure*}[t]
	\centering
	\subfloat[$\delta=0.03$]{\includegraphics[width=60mm]{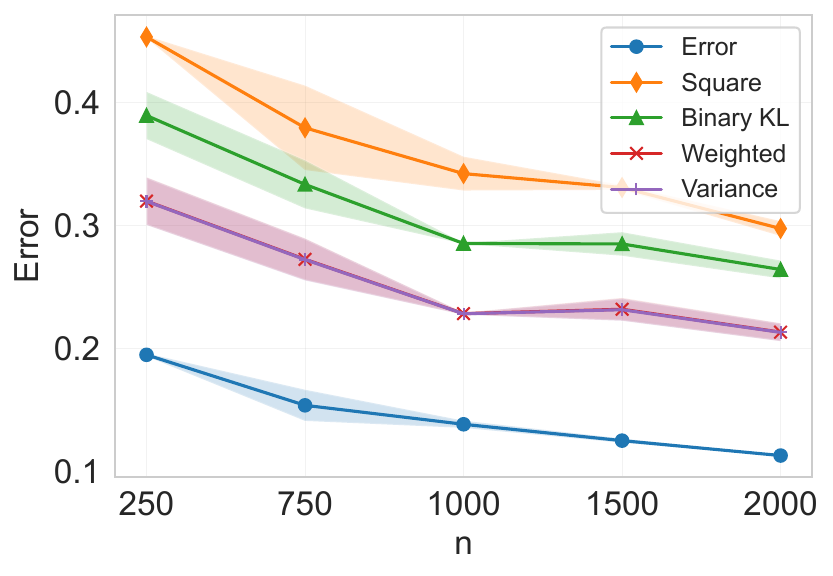}%mnist_01_8.pdf
	}
	\subfloat[$\delta=0.06$]{\includegraphics[width=60mm]{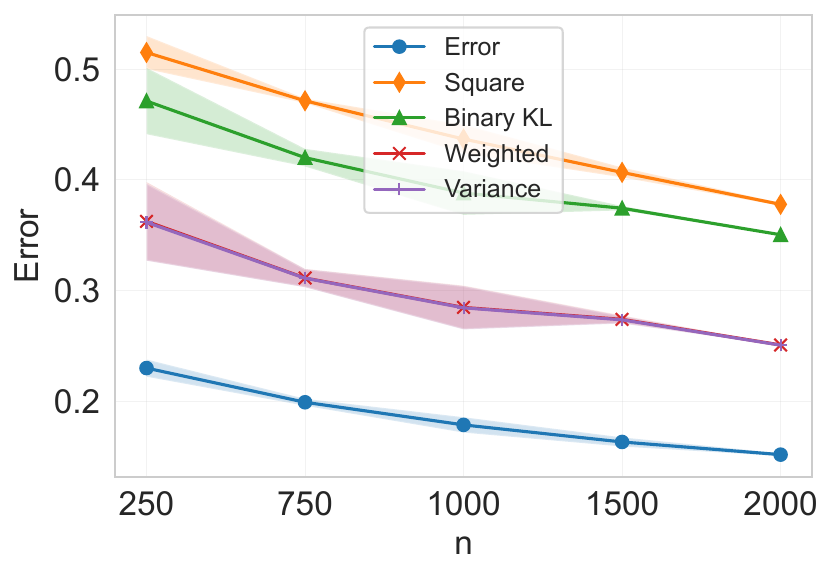}%
	}
	\subfloat[$\delta=0.09$]{\includegraphics[width=60mm]{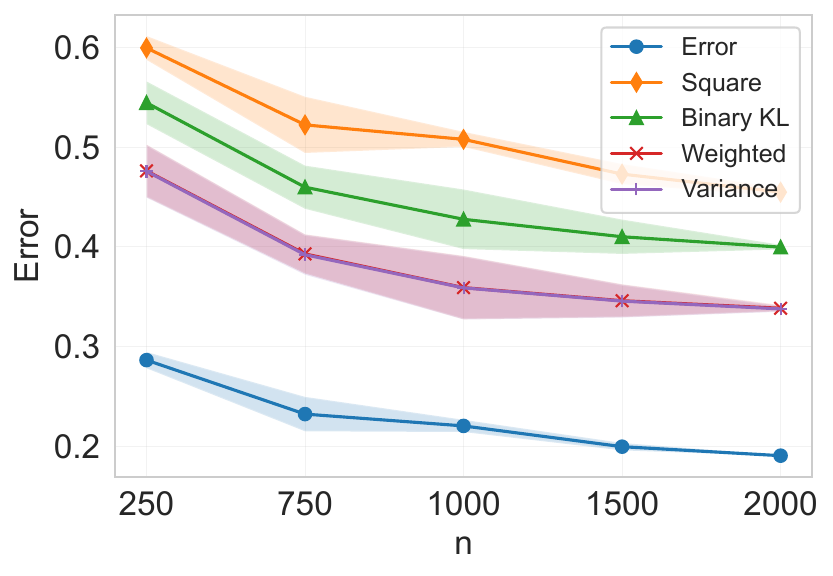}%
	}
	\caption{Comparison of the generalization error and the derived bounds for the MNIST classification task with different levels of label noise, where the labels are randomly flipped with probability $\delta$ and the memory buffer size $m=400$.}
	\label{fig_4}
\end{figure*}
In this section, we empirically evaluate the tightness of our derived bounds in previous sections, w.r.t. the true generalization errors, including the square-root (Theorem \ref{e-mi}), binary KL (Theorem \ref{binarykl2}), and fast-rate bounds (Theorems \ref{fast-loss}\&\ref{fast-var}). We employ a binary loss function to quantify empirical and population risks. All experiments are conducted with three independent runs using different random seeds, and the mean and standard deviation are reported.

\subsection{Experimental Details}
\paragraph{Datasets} We employ two benchmark image classification datasets in CL settings: the MNIST and CIFAR-10 datasets. MNIST consists of $70,000$ grayscale images of handwritten digits ranging from $0$ to $9$, while CIFAR-10 comprises $60,000$ color images across $10$ different categories. For each dataset, we construct CL problems by splitting $10$ digits/classes into $5$ sequential tasks, where each task contains $2$ distinct classes. Additionally, we construct the memory buffer for experience replay by employing a random balanced sampling strategy to select exemplars from previously encountered tasks.
\paragraph{Training Details and Estimation} Following prior work \cite{hellstrom2022new,dongtowards}, we train a four-layer convolutional neural network (CNN) on the MNIST dataset and fine-tune a pretrained ResNet-50 model on the CIFAR-10 dataset. Unless otherwise specified, model parameters are optimized using the Adam algorithm with a learning rate of $1\mathrm{e}^{-3}$. After the training process, we quantitatively estimate the derived upper bounds by computing the training and test losses on supersamples drawn from the datasets and quantifying the corresponding loss-based MI terms.

\subsection{Experimental Results}

We first investigate the generalization error and the derived bounds under varying memory buffer sizes $m$ and current task data sizes $n$. As illustrated in Figure \ref{fig_1}, our bounds exhibit a decreasing trend as $m$ or $n$ increases, aligning well with the convergence behavior of the generalization error. Among these, both the weighted and variance-based bounds consistently provide the tightest upper bound estimates, confirming our analysis on the effectiveness of the fast-rate bound under low training risk.

The subsequent experiment further assesses the scalability of the derived bounds under diverse learning algorithms, where the size of the memory buffer is fixed to $400$. As illustrated in Figure \ref{fig_2}, our bounds effectively capture the generalization dynamics across multiple deep-learning scenarios. Specifically for the SGLD algorithm, we additionally investigate the impact of key hyperparameters, specifically the learning rate $\eta$ and injected noise variance $\theta$, on the generalization guarantees. The experimental results on the MNIST dataset presented in Figure \ref{fig_3} demonstrate that smaller values of $\eta$ combined with larger $\theta$ lead to tighter upper bounds, which is consistent with our theoretical findings in Theorem \ref{algorithm}.

To examine scenarios prone to significant overfitting, we introduce random label noise into the MNIST dataset by randomly flipping the label with a specified probability $\delta$. As illustrated in Figure \ref{fig_4}, the evaluated generalization bounds consistently provide non-vacuous estimates of the generalization error. The fast-rate bounds (Theorems \ref{fast-loss} and \ref{fast-var}) emerge as the most stringent among these comparisons.

\section{Conclusion}
In this paper, we provide a unified generalization analysis for replay-based CL by developing various information-theoretic generalization error bounds. The derived bounds, expressed in terms of diverse information measures, characterize the interplay among the learning algorithm, the sequence of tasks, and the generalization error. In particular, the loss-based bounds provide tighter and more computationally tractable estimates of the generalization error. Our analysis is modular and broadly applicable to a wide range of learning algorithms. Numerical results on real-world datasets demonstrate the effectiveness of our bounds in capturing the generalization dynamics in CL settings. In future work, we will develop theory-driven CL algorithms that achieve excellent generalization performance on downstream tasks while effectively mitigating catastrophic forgetting.

\section{Acknowledgment}
This work was supported in part by the National Natural Science Foundation of China under Grants 62192781, 62576268, 62137002, the Key Research and Development Project in Shaanxi Province No. 2024PT-ZCK-89, the Project of China Knowledge Centre for Engineering Science and Technology No. 2023GXLH-024, and the Fundamental Research Funds for the Central Universities No. xxj032025002.

\bibliographystyle{IEEEtran}
\bibliography{IEEEexample}

\appendix

\section{Preparatory Definitions and Lemmas}
\begin{definition}[$\sigma$-sub-gaussian]
	A random variable $X$ is $\sigma$-sub-gaussian if for any $\lambda$, $\ln \mathbb{E}[e^{\lambda (X-\mathbb{E}X) }]\leq \lambda^2 \sigma^2/2$.
\end{definition}
\begin{definition}[Binary Relative Entropy]
	Let $p,q\in[0,1]$. Then $d(p\Vert q)$ denotes the relative entropy between two Bernoulli random variables with parameters $p$ and $q$ respectively, defined as $d(p\Vert q)=p\log(\frac{p}{q})+(1-p)\log(\frac{1-p}{1-q})$. Given $\gamma\in\mathbb{R}$, $d_\gamma(p\Vert q)=\gamma p-\log(1-q+qe^{\gamma})$ is the relaxed version of binary relative entropy. One can prove that $\sup_\gamma d_\gamma(p\Vert q)=d(p\Vert q)$.
\end{definition}
\begin{definition}[Kullback-Leibler Divergence]
	Let $P$ and $Q$ be probability distributions defined on the same measurable space such that $P$ is absolutely continuous with respect to $Q$. The Kullback-Leibler (KL) divergence between $P$ and $Q$ is defined as $D(P\Vert Q)\triangleq \int_{\mathcal{X}}p(x)\log (\frac{p(x)}{q(x)})dx$.
\end{definition}
\begin{definition}[Mutual Information]
	For random variables $X$ and $Y$ with joint distribution $P_{X,Y}$ and product of their marginals $P_XP_Y$, the mutual information between $X$ and $Y$ is defined as $I(X;Y) = D(P_{X,Y} \Vert P_XP_Y)$.
\end{definition}
\begin{lemma}[Donsker-Varadhan Formula (Theorem 5.2.1 in \cite{gray2011entropy})] \label{lemmaA.5}
	Let $P$ and $Q$ be probability measures over the same space $\mathcal{X}$ such that $P$ is absolutely continuous with respect to $Q$. For any bounded function $f:\mathcal{X}\rightarrow \mathbb{R}$, 
	\begin{equation*}
		D(P\Vert Q) = \sup_{f} \Big\{ \mathbb{E}_{X\sim P}[f(X)] - \log \mathbb{E}_{X\sim Q} [e^{f(X)}] \Big\},
	\end{equation*}
	where $X$ is any random variable such that both $\mathbb{E}_{X\sim P}[f(X)]$ and $\mathbb{E}_{X\sim Q} [e^{f(X)}]$ exist.
\end{lemma}
\begin{lemma}[Lemma 1 in \cite{harutyunyan2021information}]\label{lemmaA.6}
	Let $(X,Y)$ be a pair of random variables with joint distribution $P_{X,Y}$, and $\bar{Y}$ be an independent copy of $Y$. If $f(x,y)$ be a measurable function such that $\mathbb{E}_{X,Y}[f(X,Y)]$ exists and $\mathbb{E}_{X,\bar{Y}}[f(X,\bar{Y})]$ is $\sigma$-sub-gaussian, then 
	\begin{equation*}
		\big\vert \mathbb{E}_{X,Y} [f(X,Y)] - \mathbb{E}_{X,\bar{Y}}[f(X,\bar{Y})] \big\vert \leq \sqrt{2\sigma^2 I(X,Y)}.
	\end{equation*}
\end{lemma}
\begin{lemma}[Lemma 3 in \cite{harutyunyan2021information}] \label{lemmaA.7}
	Let $X$ and $Y$ be independent random variables. If $f$ is a measurable function such that $f(x,Y)$ is $\sigma$-sub-gaussian and $\mathbb{E}f(x,Y)=0$ for all $x\in\mathcal{X}$, then $f(X,Y)$ is also $\sigma$-sub-gaussian.
\end{lemma}
\begin{lemma}[Lemma 2 in \cite{hellstrom2022evaluated}] \label{lemmaA.8}
	Let $X_1,\ldots,X_n$ be $n$ independent random variables that for $i\in[n]$, $X_i\sim P_{X_i}$, $\mathbb{E}[X_i]=\mu_i$, $\bar{\mu}=\frac{1}{n}\sum_{i=1}^{n}X_i$, and $\mu = \frac{1}{n}\sum_{i=1}^{n}\mu_i$. Assume that $X_i\in[0,1]$ almost surely. Then, for any $\gamma>0$, $\mathbb{E}[e^{(n d_\gamma(\bar{\mu} \Vert \mu))}]\leq 1$.
\end{lemma}
\begin{lemma}[Lemma A.11 in \cite{dongtowards}] \label{lemmaA.9}
	Let $X\sim N(0,\Sigma)$ and $Y$ be any zero-mean random vector satisfying $\mathrm{Cov}_Y[Y] = \Sigma$, then $H(Y)\leq H(X)$.
\end{lemma}
\begin{lemma}[Lemma 9 in \cite{dong2023understanding}] \label{lemmaA.10}
	For any symmetric positive-definite matrix $A$, let $A = \Big[\begin{matrix}  B & D^T \\ D & C \end{matrix}\Big] $ be a partition of $A$, where $B$ and $C$ are square matrices, then $\vert A \vert \leq \vert B \vert \vert C \vert$.
\end{lemma}

\section{Detailed Proofs in Section \ref{Section3}}
\subsection{Proof of Theorem \ref{input-output}}
\begin{restatetheorem}{\ref{input-output}}[Restate]
	Let $n$ and $\tilde{n}$ denote the number of samples available for training the current task $D^t$ and the number of samples from the previous task $i$ in memory $M^i$, respectively, where $t\in[T]$ and $i\in[t-1]$. Assume that $\ell(w,Z)$, where $Z\in\mathcal{Z}$, is $\sigma$-subgaussian for all $w\in\mathcal{W}$, we have 
	\begin{equation*}
		\vert \mathrm{gen}_{W} \vert \leq   \underbrace{\sqrt{\frac{2\sigma^2}{n} I(W;D^t)}}_{\text{Current Task Generalization}} + \sum_{i=1}^{t-1} \bigg(\underbrace{\sqrt{\frac{2\sigma^2}{n} I(W;M^i)}}_{\text{Previous Task Generalization}} + \underbrace{\sqrt{\frac{2\sigma^2(n-\tilde{n})}{n\tilde{n}} I(W;M^i)}}_{\text{Memory Compression Cost}}\bigg).
	\end{equation*}   
\end{restatetheorem}

\begin{proof}
	According to the definition of generalization error, we have
	\begin{align}
		\vert \mathrm{gen}_{W} \vert = & \Bigg\vert \mathbb{E}_{W,D^{1:T}}\Bigg[\sum_{i=1}^t \mathbb{E}_{Z\sim \mu_i}[\ell(W,Z)] - \bigg(\sum_{i=1}^{t-1} \frac{1}{\tilde{n}}\sum_{j=1}^{\tilde{n}} \ell(W,Z^{i}_{j}) + \frac{1}{n}\sum_{i=1}^n \ell(W,Z^{t}_{j})\bigg) \Bigg] \Bigg\vert \nonumber\\
		= & \Bigg\vert\mathbb{E}_{W,D^{1:T}} \Bigg[\sum_{i=1}^t \mathbb{E}_{Z\sim \mu_i}[\ell(W,Z)] - \Bigg(\sum_{i=1}^{t-1} \frac{1}{\tilde{n}}\sum_{j=1}^{\tilde{n}} \ell(W,Z^{i}_{j}) + \frac{1}{n}\sum_{i=1}^n \ell(W,Z^{t}_{j})\Bigg) \nonumber\\
		& + \sum_{i=1}^{t-1} \frac{1}{n}\sum_{j=1}^{n} \ell(W,Z^{i}_{j}) -  \sum_{i=1}^{t-1} \frac{1}{n}\sum_{j=1}^{n} \ell(W,Z^{i}_{j}) \Bigg]\Bigg\vert\nonumber\\
		\leq & \Bigg\vert \mathbb{E}_{W,D^{1:T}} \Bigg[ \sum_{i=1}^t \left(\mathbb{E}_{Z\sim \mu_i}[\ell(W,Z)] - \frac{1}{n}\sum_{j=1}^{n} \ell(W,Z^{i}_{j}) \right) \Bigg] \Bigg\vert + \Bigg\vert\mathbb{E}_{W,D^{1:T}}\Bigg[ \sum_{i=1}^{t-1} \left( \frac{1}{n}\sum_{j=1}^{n} \ell(W,Z^{i}_{j}) - \frac{1}{\tilde{n}}\sum_{j=1}^{\tilde{n}} \ell(W,Z^{i}_{j})\right)\Bigg] \Bigg\vert\nonumber\\
		= & \Bigg\vert  \mathbb{E}_{W,D^{1:T}} \Bigg[ \sum_{i=1}^t \left(\mathbb{E}_{Z\sim \mu_i}[\ell(W,Z)] - \frac{1}{n}\sum_{j=1}^{n} \ell(W,Z^{i}_{j}) \right) \Bigg] \Bigg\vert  \nonumber\\
		& + \Bigg\vert \mathbb{E}_{W,D^{1:T}}\Bigg[ \sum_{i=1}^{t-1} \left( \frac{\tilde{n}}{n}\times \frac{1}{\tilde{n}}\sum_{j=1}^{\tilde{n}} \ell(W,Z^{i}_{j}) +\frac{n-\tilde{n}} {n}\times \frac{1}{n-\tilde{n}}\sum_{j=\tilde{n}+1}^{n} \ell(W,Z^{i}_{j})   - \frac{1}{\tilde{n}}\sum_{j=1}^{\tilde{n}} \ell(W,Z^{i}_{j})\right)\Bigg] \Bigg\vert  \nonumber\\
		= & \Bigg\vert \mathbb{E}_{W,D^{1:T}} \Bigg[ \sum_{i=1}^t \left(\mathbb{E}_{Z\sim \mu_i}[\ell(W,Z)] - \frac{1}{n}\sum_{j=1}^{n} \ell(W,Z^{i}_{j}) \right) \Bigg] \Bigg\vert \nonumber\\
		& + \Bigg\vert  \mathbb{E}_{W,D^{1:T}}\Bigg[ \sum_{i=1}^{t-1} \left( \frac{\tilde{n}}{n}\times \frac{1}{\tilde{n}}\sum_{j=1}^{\tilde{n}} \ell(W,Z^{i}_{j}) +\frac{n-\tilde{n}} {n}\times \frac{1}{n-\tilde{n}}\sum_{j=\tilde{n}+1}^{n} \ell(W,Z^{i}_{j})   - \frac{1}{\tilde{n}}\sum_{j=1}^{\tilde{n}} \ell(W,Z^{i}_{j})\right)\Bigg] \Bigg\vert  \nonumber\\
		\leq & \sum_{i=1}^t \Bigg\vert   \mathbb{E}_{W,D^{i}} \bigg[\bigg(\mathbb{E}_{Z\sim \mu_i}[\ell(W,Z)] - \frac{1}{n}\sum_{j=1}^{n} \ell(W,Z^{i}_{j}) \bigg) \bigg]  \Bigg\vert \nonumber\\
		& +  \frac{n-\tilde{n}}{n} \sum_{i=1}^{t-1} \Bigg\vert  \mathbb{E}_{W,D^{i}} \bigg[ \frac{1}{n-\tilde{n}}\sum_{j=\tilde{n}+1}^{n} \ell(W,Z^{i}_{j}) - \frac{1}{\tilde{n}}\sum_{j=1}^{\tilde{n}} \ell(W,Z^{i}_{j}) \bigg]\Bigg\vert  \label{eqq1}
	\end{align}
	We proceed to bound the two terms on the right side of equation (\ref{eqq1}), respectively. For the first term of (\ref{eqq1}), let \begin{equation*}
		f(W,D^i) = \frac{1}{n}\sum_{j=1}^n \ell(W,Z^{i}_{j}),
	\end{equation*}
	and $\bar{D}^i=\{\bar{Z}^i_j\}_{j=1}^n$ is an independent copy of $D^i$ for all $i\in[t]$. Applying Donsker-Varadhan inequality in Lemma \ref{lemmaA.5} with $P=P_{W,D^i}$, $Q=P_WP_{D^i}$, and $f(X,Y) = f(W,D^i)$ where $X=W$ and $Y=D^i$, we get 
	\begin{align}
		I(W;D^i) = & D(P_{W,D^i} \Vert P_WP_{D^i}) \nonumber\\
		\geq & \sup_{\lambda} \Big\{\lambda\Big( \mathbb{E}_{W,D^i} \big[f(W,D^i)\big]  - \mathbb{E}_{W,\bar{D}^i } \big[f(W,\bar{D}^i)\big]\Big) - \log \mathbb{E}_{W,\bar{D}^i} \Big[ e^{\lambda (f(W,\bar{D}^i)-\mathbb{E}[f(W,\bar{D}^i)]) }\Big] \Big\} .  \label{eq1}
	\end{align}
	By the subgaussian property of the loss function, it is clear that $f(W,\bar{D}^i)$ is $\frac{\sigma}{\sqrt{n}}$-subgaussian random variable, which implies that
	\begin{align*}
		\log \mathbb{E}_{W,\bar{D}^i} \Big[ e^{\lambda (f(W,\bar{D}^i)-\mathbb{E}[f(W,\bar{D}^i)]) }\Big] \leq \frac{\lambda^2 \sigma^2}{2n}.
	\end{align*}
	Putting the above back into the inequality (\ref{eq1}), we have
	\begin{align*}
		I(W;D^i)
		\geq \sup_{\lambda} \Big\{\lambda\Big( \mathbb{E}_{W,D^i} \big[f(W,D^i)\big]  - \mathbb{E}_{W,\bar{D}^i } \big[f(W,\bar{D}^i)\big]\Big)  -  \frac{\lambda^2 \sigma^2}{2n}\Big\} 
	\end{align*}
	Solving $\lambda$ to maximize the RHS of the above inequality, we get that
	\begin{align*}
		\bigg\vert  \mathbb{E}_{W,D^i} \big[f(W,D^i)\big]  - \mathbb{E}_{W,\bar{D}^i } \big[f(W,\bar{D}^i)\big] \bigg\vert 
		=&\bigg\vert \mathbb{E}_{W,D^i} \Big[ \frac{1}{n}\sum_{j=1}^n \ell(W,Z^{i}_{j})\Big]  - \mathbb{E}_{W,\bar{D}^i }\Big[\frac{1}{n}\sum_{j=1}^n \ell(W,\bar{Z}^{i}_{j})\Big] \bigg\vert \\
		= & \bigg\vert \mathbb{E}_{W,D^i} \Big[ \frac{1}{n}\sum_{j=1}^n \ell(W,Z^{i}_{j})\Big]  - \mathbb{E}_{W,D^i}\Big[\mathbb{E}_{Z\sim\mu_i} \ell(W,\bar{Z}^{i}_{j})\Big] \bigg\vert \\
		\leq & \sqrt{\frac{2\sigma^2}{n} I(W;D^i)}.
	\end{align*}
	Putting the above inequality back into the first term on the right side of (\ref{eqq1}), we obtain that
	\begin{equation}\label{decom1}
		\sum_{i=1}^t   \mathbb{E}_{W,D^{i}} \bigg[\bigg(\mathbb{E}_{Z\sim \mu_i}[\ell(W,Z)] - \frac{1}{n}\sum_{j=1}^{n} \ell(W,Z^{i}_{j}) \bigg) \bigg] \leq \sum_{i=1}^t \sqrt{\frac{2\sigma^2}{n} I(W;D^i)}= \sum_{i=1}^{t-1} \sqrt{\frac{2\sigma^2}{n} I(W;M^i)} +  \sqrt{\frac{2\sigma^2}{n} I(W;D^t)},
	\end{equation}
	where the last equality follows from the fact that $W$ is trained using only data $M^i$ for $i\in[t-1]$, and thus $I(W;D^i\backslash M^i|M^i)=0$ and $I(W;D^i) = I(W;M^i) + I(W;D^i\backslash M^i|M^i)=I(W;M^i)$.
	We then bound the second term on the right side of (\ref{eqq1}). For all $i\in[t-1]$, let 
	\begin{equation*}
		\hat{f}(W,D^i) = \frac{1}{n-\tilde{n}}\sum_{j=\tilde{n}+1}^{n} \ell(W,Z^{i}_{j}) - \frac{1}{\tilde{n}}\sum_{j=1}^{\tilde{n}} \ell(W,Z^{i}_{j}).
	\end{equation*}
	Let $\bar{D}^i=\{\bar{Z}^i_j\}_{j=1}^n$ is an independent copy of $D^i$. By using Donsker-Varadhan inequality in Lemma \ref{lemmaA.5} with $P=P_{W,D^i}$, $Q=P_WP_{D^i}$, and $f(X,Y) = \hat{f}(W,D^i)$, for all $i\in[t-1]$, we have 
	\begin{align}
		I(W;D^i) \geq & \sup_{\lambda} \Big\{\lambda\Big( \mathbb{E}_{W,D^i} \big[\hat{f}(W,D^i)\big]  - \mathbb{E}_{W,\bar{D}^i } \big[\hat{f}(W,\bar{D}^i)\big]\Big) - \log \mathbb{E}_{W,\bar{D}^i} \Big[ e^{\lambda (\hat{f}(W,\bar{D}^i)-\mathbb{E}[\hat{f}(W,\bar{D}^i)]) }\Big] \Big\} .  \label{eq2}
	\end{align}
	Notice that if the loss function $\ell$ is $\sigma$-subGaussian, one can prove that $\hat{f}(W,\bar{D}^i)$ is $\sigma\sqrt{\frac{1}{n-\tilde{n}}+\frac{1}{\tilde{n}}}$-subGaussian random variable. We then have 
	\begin{equation*}
		\log \mathbb{E}_{W,\bar{D}^i} \Big[ e^{\lambda (\hat{f}(W,\bar{D}^i)-\mathbb{E}[\hat{f}(W,\bar{D}^i)]) }\Big] \leq \frac{\lambda^2 \sigma^2}{2} (\frac{1}{n-\tilde{n}}+\frac{1}{\tilde{n}}).
	\end{equation*}
	Substituting the above into inequality (\ref{eq2}) and solving $\lambda$, we get 
	\begin{equation*}
		\Big\vert  \mathbb{E}_{W,D^i} \big[\hat{f}(W,D^i)\big]  - \mathbb{E}_{W,\bar{D}^i } \big[\hat{f}(W,\bar{D}^i)\big] \Big\vert \leq \sqrt{2\sigma^2(\frac{1}{n-\tilde{n}}+\frac{1}{\tilde{n}}) I(W;D^i)}.
	\end{equation*}
	Note that $\mathbb{E}_{\bar{D}_i}[\hat{f}(w,D^i)]=0$ for each $w$, we thus have 
	\begin{equation}\label{eqq2}
		\Big\vert \mathbb{E}_{W,D^i}\Big[\frac{1}{n-\tilde{n}}\sum_{j=\tilde{n}+1}^{n} \ell(W,Z^{i}_{j}) - \frac{1}{\tilde{n}}\sum_{j=1}^{\tilde{n}} \ell(W,Z^{i}_{j}) \Big] \Big\vert \leq \sqrt{2\sigma^2(\frac{1}{n-\tilde{n}}+\frac{1}{\tilde{n}}) I(W;D^i)}.
	\end{equation}
	Plugging the inequality (\ref{eqq2}) into the second term on the right side of (\ref{eqq1}), we obtain
	\begin{align}
		\frac{n-\tilde{n}}{n} \sum_{i=1}^{t-1} \Bigg\vert  \mathbb{E}_{W,D^{i}} \bigg[ \frac{1}{n-\tilde{n}}\sum_{j=\tilde{n}+1}^{n} \ell(W,Z^{i}_{j}) - \frac{1}{\tilde{n}}\sum_{j=1}^{\tilde{n}} \ell(W,Z^{i}_{j}) \bigg]\Bigg\vert  \leq & \sum_{i=1}^{t-1} \sqrt{\frac{2\sigma^2(n-\tilde{n})}{n\tilde{n}} I(W;D^i)} \nonumber\\
		= & \sum_{i=1}^{t-1} \sqrt{\frac{2\sigma^2(n-\tilde{n})}{n\tilde{n}} I(W;M^i)}, \label{decom2}
	\end{align}
	where the last equality follows from the fact that $W$ is trained using only data $M^i$ for $i\in[t-1]$, and thus $I(W;D^i\backslash M^i|M^i)=0$ and $I(W;D^i) = I(W;M^i) + I(W;D^i\backslash M^i|M^i)=I(W;M^i)$. Substituting the inequalities (\ref{decom1}) and (\ref{decom2}) into (\ref{eqq1}), we have 
	\begin{equation*}
		\vert \mathrm{gen}_{W} \vert \leq  \sqrt{\frac{2\sigma^2}{n} I(W;D^t)} + \sum_{i=1}^{t-1} \bigg(\sqrt{\frac{2\sigma^2}{n} I(W;M^i)} +\sqrt{\frac{2\sigma^2(n-\tilde{n})}{n\tilde{n}} I(W;M^i)}\bigg).
	\end{equation*}
	This completes the proof.
\end{proof}
\subsection{Proof of Theorem \ref{cmi}}
\begin{restatetheorem}{\ref{cmi}}[Restate]
	Let $n$ and $\tilde{n}$ denote the number of samples available for training the current task and the number of the memory, respectively. Assume that $\ell(w,Z)$, where $Z\in\mathcal{Z}$, is $\sigma$-subgaussian for all $w\in\mathcal{W}$, we have 
	\begin{equation*}
		\vert \mathrm{gen}_{W} \vert \leq \underbrace{\mathbb{E}_{\tilde{D}^{t}} \sqrt{\frac{4\sigma^2 I^{\tilde{D}^t}(W; S^i)}{n}}}_{\text{Current Task Generalization}} + \underbrace{\mathbb{E}_{\tilde{D}^{1:t-1}} \sum_{i=1}^{t-1} \sqrt{\frac{4\sigma^2I^{\tilde{D}^i}(W; S^i)}{n}}}_{\text{Previous Task Generalization}} + \underbrace{\mathbb{E}_{\tilde{D}^{1:t-1},S^{1:t-1}} \sum_{i=1}^{t-1}\sqrt{\frac{2\sigma^2 (n-\tilde{n})}{n\tilde{n}} I^{\tilde{D}^{i}_{S^i}}(W; U^{i})}}_{\text{Memory Compression Cost}}.
	\end{equation*}
\end{restatetheorem}
\begin{proof}
	By leveraging supersample settings and the definition of generalization error, we have
	\begin{align}
		\vert \mathrm{gen}_{W} \vert 
		= & \Bigg\vert \mathbb{E}_{W,\tilde{D}^{1:t},S^{1:t}, U^{1:t-1}}\Bigg[\sum_{i=1}^t \frac{1}{n} \sum_{j=1}^{n} \ell(W,\tilde{Z}^i_{j,1-S^i_j}) - \bigg(\sum_{i=1}^{t-1} \frac{1}{\tilde{n}}\sum_{j=1}^{n} U^i_j \ell(W,\tilde{Z}^i_{j,S^i_j}) + \frac{1}{n}\sum_{i=1}^n \ell(W,\tilde{Z}^i_{j,S^i_j})\bigg)  \nonumber\\
		& + \sum_{i=1}^{t-1} \frac{1}{n} \sum_{j=1}^{n} \ell(W,\tilde{Z}^i_{j,S^i_j}) -  \sum_{i=1}^{t-1} \frac{1}{n} \sum_{j=1}^{n} \ell(W,\tilde{Z}^i_{j,S^i_j}) \Bigg]\Bigg\vert\nonumber\\
		\leq & \Bigg\vert \mathbb{E}_{W,\tilde{D}^{1:t},S^{1:t}}\Bigg[\sum_{i=1}^t \frac{1}{n} \sum_{j=1}^{n} \Big(\ell(W,\tilde{Z}^i_{j,1-S^i_j}) - \ell(W,\tilde{Z}^i_{j,S^i_j})\Big) \Bigg] \Bigg\vert \nonumber\\
		&  +  \Bigg\vert \mathbb{E}_{W,\tilde{D}^{1:t},S^{1:t}, U^{1:t-1}}\Bigg[\sum_{i=1}^{t-1} \Big( \frac{1}{n} \sum_{j=1}^{n} \ell(W,\tilde{Z}^i_{j,1-S^i_j}) -  \frac{1}{\tilde{n}}\sum_{j=1}^{n} U^i_j \ell(W,\tilde{Z}^i_{j,S^i_j})\Big) \Bigg] \Bigg\vert \nonumber \\
		\leq & \mathbb{E}_{\tilde{D}^{1:t}} \Bigg\vert \mathbb{E}_{W,S^{1:t}|\tilde{D}^{1:t}}\Bigg[ \sum_{i=1}^t  \frac{1}{n} \sum_{j=1}^{n} \Big(\ell(W,\tilde{Z}^i_{j,1-S^i_j}) - \ell(W,\tilde{Z}^i_{j,S^i_j})\Big) \Bigg] \Bigg\vert \nonumber \\
		&  + \mathbb{E}_{\tilde{D}^{1:t-1},S^{1:t-1}} \Bigg\vert \mathbb{E}_{W, U^{1:t-1}|\tilde{D}^{1:t-1},S^{1:t-1}}\Bigg[\sum_{i=1}^{t-1} \Big( \frac{1}{n} \sum_{j=1}^{n} \ell(W,\tilde{Z}^i_{j,1-S^i_j}) -  \frac{1}{\tilde{n}}\sum_{j=1}^{n} U^i_j \ell(W,\tilde{Z}^i_{j,S^i_j})\Big) \Bigg] \Bigg\vert \nonumber \\
		\leq & \mathbb{E}_{\tilde{D}^{1:t}} \sum_{i=1}^t  \Bigg\vert \mathbb{E}_{W,S^i|\tilde{D}i}\Bigg[  \frac{1}{n} \sum_{j=1}^{n} \Big(\ell(W,\tilde{Z}^i_{j,1-S^i_j}) - \ell(W,\tilde{Z}^i_{j,S^i_j})\Big) \Bigg] \Bigg\vert \nonumber \\
		&  + \mathbb{E}_{\tilde{D}^{1:t-1},S^{1:t-1}} \sum_{i=1}^{t-1} \Bigg\vert \mathbb{E}_{W, U^{i}|\tilde{D}^{i}_{S^i}}\Bigg[ \frac{1}{n} \sum_{j=1}^{n} \ell(W,\tilde{Z}^i_{j,1-S^i_j}) -  \frac{1}{\tilde{n}}\sum_{j=1}^{n} U^i_j \ell(W,\tilde{Z}^i_{j,S^i_j}) \Bigg] \Bigg\vert.  \label{ineq:9} 
	\end{align}
	For the first term of inequality (\ref{ineq:9}), let $f(W,S^i) = \frac{1}{n} \sum_{j=1}^{n} \big(\ell(W,\tilde{Z}^i_{j,1-S^i_j}) - \ell(W,\tilde{Z}^i_{j,S^i_j})\big)$. For each value of $w$, it can be proven that $f(w,\bar{S}^i)$ is a $\sigma\sqrt{2/n}$-subGaussian random variable, and $\forall w$, $\mathbb{E}_{\bar{S}^i}[f(w,\bar{S}^i)]=0$. Leveraging Lemma \ref{lemmaA.7} and the above statements, this implies that $f(W,\bar{S}^i)$ is also $\sigma\sqrt{2/n}$-subGaussian. By applying Lemma \ref{lemmaA.6}, we further have 
	\begin{align}
		& \Bigg\vert \mathbb{E}_{W,S^i|\tilde{D}^i}\Bigg[  \frac{1}{n} \sum_{j=1}^{n} \Big(\ell(W,\tilde{Z}^i_{j,1-S^i_j}) - \ell(W,\tilde{Z}^i_{j,S^i_j})\Big) \Bigg] - \mathbb{E}_{W,\bar{S}^i|\tilde{D}^i}\Bigg[  \frac{1}{n} \sum_{j=1}^{n} \Big(\ell(W,\tilde{Z}^i_{j,1-\bar{S}^i_j}) - \ell(W,\tilde{Z}^i_{j,\bar{S}^i_j})\Big) \Bigg] \Bigg\vert \nonumber\\
		= &\Bigg\vert \mathbb{E}_{W,S^i|\tilde{D}^i}\Bigg[  \frac{1}{n} \sum_{j=1}^{n} \Big(\ell(W,\tilde{Z}^i_{j,1-S^i_j}) - \ell(W,\tilde{Z}^i_{j,S^i_j})\Big) \Bigg]   \Bigg\vert \nonumber\\
		\leq & 2\sigma\sqrt{\frac{I^{\tilde{D}^i}(W; S^i)}{n}}.  \label{ineq:10}
	\end{align}
	Putting the inequality (\ref{ineq:10}) back into the first term of inequality (\ref{ineq:9}), we have the following upper bound:
	\begin{equation}\label{ineq:11}
		\mathbb{E}_{\tilde{D}^{1:t}} \sum_{i=1}^t  \Bigg\vert \mathbb{E}_{W,S^i|\tilde{D}i}\Bigg[  \frac{1}{n} \sum_{j=1}^{n} \Big(\ell(W,\tilde{Z}^i_{j,1-S^i_j}) - \ell(W,\tilde{Z}^i_{j,S^i_j})\Big) \Bigg] \Bigg\vert  \leq  \mathbb{E}_{\tilde{D}^{1:t}} 2\sigma \sum_{i=1}^t \sqrt{\frac{I^{\tilde{D}^i}(W; S^i)}{n}}.
	\end{equation}
	Similarly, for the second term of inequality (\ref{ineq:9}), let $f(W,U^i)= \frac{1}{n} \sum_{j=1}^{n} \ell(W,\tilde{Z}^i_{j,1-S^i_j}) -  \frac{1}{\tilde{n}}\sum_{j=1}^{n} U^i_j \ell(W,\tilde{Z}^i_{j,S^i_j})$. Note that for each $w$, it is proven that the random variable $f(w,\bar{U}^i)$ is $\sigma\sqrt{\frac{1}{\tilde{n}} - \frac{1}{n}}$-subGaussian. Furthermore, $\forall w$, $\mathbb{E}_{\bar{U}^i}[f(w,\bar{U}^i)]=0$. These two statements together and Lemma \ref{lemmaA.7} imply that $f(W, \bar{U}^i)$ is also $\sigma\sqrt{\frac{1}{\tilde{n}} - \frac{1}{n}}$-subGaussian. Leveraging Lemma \ref{lemmaA.6} implies that 
	\begin{equation}
		\Bigg\vert \mathbb{E}_{W, U^{i}|\tilde{D}^{i}_{S^i}}\Bigg[ \frac{1}{n} \sum_{j=1}^{n} \ell(W,\tilde{Z}^i_{j,1-S^i_j}) -  \frac{1}{\tilde{n}}\sum_{j=1}^{n} U^i_j \ell(W,\tilde{Z}^i_{j,S^i_j}) \Bigg] \Bigg\vert \leq \sqrt{\frac{2\sigma^2 (n-\tilde{n})}{n\tilde{n}} I^{\tilde{D}^{i}_{S^i}}(W; U^{i})}.
	\end{equation}
	Plugging the above inequality into the second term of inequality (\ref{ineq:9}), we have 
	\begin{align}
		& \mathbb{E}_{\tilde{D}^{1:t-1},S^{1:t-1}} \sum_{i=1}^{t-1} \Bigg\vert \mathbb{E}_{W, U^{i}|\tilde{D}^{i}_{S^i}}\Bigg[ \frac{1}{n} \sum_{j=1}^{n} \ell(W,\tilde{Z}^i_{j,1-S^i_j}) -  \frac{1}{\tilde{n}}\sum_{j=1}^{n} U^i_j \ell(W,\tilde{Z}^i_{j,S^i_j}) \Bigg] \Bigg\vert \nonumber\\
		\leq & \mathbb{E}_{\tilde{D}^{1:t-1},S^{1:t-1}} \sum_{i=1}^{t-1}\sqrt{\frac{2\sigma^2 (n-\tilde{n})}{n\tilde{n}} I^{\tilde{D}^{i}_{S^i}}(W; U^{i})}. \label{ineq:13}
	\end{align}
	Substituting inequalities (\ref{ineq:11}) and (\ref{ineq:13}) into (\ref{ineq:9}), we obtain
	\begin{equation*}
		\vert \mathrm{gen}_{W} \vert \leq \mathbb{E}_{\tilde{D}^{t}} \sqrt{\frac{4\sigma^2 I^{\tilde{D}^t}(W; S^i)}{n}} + \mathbb{E}_{\tilde{D}^{1:t-1}} \sum_{i=1}^{t-1} \sqrt{\frac{4\sigma^2I^{\tilde{D}^i}(W; S^i)}{n}} + \mathbb{E}_{\tilde{D}^{1:t-1},S^{1:t-1}} \sum_{i=1}^{t-1}\sqrt{\frac{2\sigma^2 (n-\tilde{n})}{n\tilde{n}} I^{\tilde{D}^{i}_{S^i}}(W; U^{i})},
	\end{equation*}
	which completes the proof.
\end{proof}

\begin{restatetheorem}{\ref{algorithm}}[Restate]
	Let $n$ and $\tilde{n}$ denote the number of samples available for training the current task and the number of the memory, respectively. Assume that $\ell(w,Z)$, where $Z\in\mathcal{Z}$, is $\sigma$-subgaussian for all $w\in\mathcal{W}$. Let $W$ be the output of the SGLD algorithm after $R$ iterations at time $t$, then
	\begin{equation*}
		\vert \mathrm{gen}_{W} \vert \leq    \sqrt{\frac{\sigma^2}{n} \sum_{r=1}^{R} \log \Big\vert \frac{\eta_r^2}{\xi^2_r} \mathbb{E}_{W_{r-1}}\big[ \Sigma^r_t \big]  + I_d \Big\vert} + \sum_{i=1}^{t-1} \sqrt{\frac{2\sigma^2}{\tilde{n}} \sum_{r=1}^{R}\log \Big\vert \frac{\eta_r^2}{\xi^2_r} \mathbb{E}_{W_{r-1}}\big[ \tilde{\Sigma}^r_i \big]  + I_d \Big\vert},
	\end{equation*} 
	where $\tilde{\Sigma}^r_i = \mathrm{Cov}_{\tilde{B}^r_i}[\tilde{g}_i(W_r)]$ for all $i\in[t-1]$, and $\Sigma^r_t = \mathrm{Cov}_{B^r_t}[g_t(W_r)]$.
\end{restatetheorem}

\begin{proof}
	Applying the data-processing inequality on the Markov chain $(M^{1:t-1},D^{t})\rightarrow \Big(W_{R-1}, \eta_R \big(\sum_{i=1}^{t-1}\tilde{g}_i(W_R) + g_t(W_R) \big)+N_R \Big)\rightarrow W_{R-1} + \eta_R \big(\sum_{i=1}^{t-1}\tilde{g}_i(W_R) + g_t(W_R) \big)+N_R $, we have 
	\begin{align*}
		I(W_R;D^t) = & I\Big(W_{R-1} + \eta_R \big(\sum_{i=1}^{t-1}\tilde{g}_i(W_R) + g_t(W_R) \big)+N_R ;D^t \Big) \\
		= & I(W_{R-1} + \eta_Rg_t(W_R) +N_R ;D^t ) \\
		\leq & I(W_{R-1} , \eta_R g_t(W_R) +N_R ; D^t) \\
		= & I(W_{R-1}; D^t) + I(\eta_R g_t(W_R) +N_R ; D^t| W_{R-1}) \\
		& \cdots \\
		\leq & \sum_{r=1}^{R} I(\eta_r g_t(W_r) +N_r ; D^t| W_{r-1}).
	\end{align*}
	Notice that $N_r$ is independent of $D^t$ and $B^r_t$, we thus have 
	\begin{equation*}
		\mathrm{Cov}_{B^r_t, N_r} [\eta_r g_t(W_r) +N_r ] = \mathrm{Cov}_{B^r_t}[\eta_r g_t(W_r)] +  \mathrm{Cov}_{N_r} [N_r] =  \eta_r^2 \Sigma^r_t + \xi^2_r I_d.
	\end{equation*}
	Leveraging Lemma \ref{lemmaA.9} with $\Sigma = \eta_r^2 \Sigma^r_t + \xi^2_r I_d$ implies that  
	\begin{align*}
		& I^{W_{r-1}}(\eta_r g_t(W_r) +N_r ; D^t) \\
		=& H(\eta_r g_t(W_r) +N_r |W_{r-1}) - H(\eta_r g_t(W_r) +N_r |D^t,W_{r-1}) \\
		\leq & H(\eta_r g_t(W_r) +N_r |W_{r-1}) - H(\eta_r g_t(W_r) +N_r |B^r_t,W_{r-1}) \\
		= &  H(\eta_r g_t(W_r) +N_r  |W_{r-1}) - H(N_r) \\
		\leq & \frac{d}{2} \log(2\pi e) + \frac{1}{2} \log \Big\vert \eta_r^2 \Sigma^r_t + \xi^2_r I_d \Big\vert - \frac{d}{2} \log (2\pi e \xi^2_r) \\
		\leq & \frac{1}{2} \log \Big\vert \frac{\eta_r^2}{\xi^2_r} \Sigma^r_t + I_d \Big\vert .
	\end{align*}
	Combining the above inequality and employing Jensen's inequality, we obtain
	\begin{align}
		I(W_R;D^t)\leq & \sum_{r=1}^{R} I(\eta_r g_t(W_r) +N_r ; D^t| W_{r-1}) \nonumber\\
		= &\sum_{r=1}^{R} \mathbb{E}_{W_{r-1}}\Big[ I^{W_{r-1}}(\eta_r g_t(W_r) +N_r ; D^t)\Big] \nonumber\\
		\leq & \sum_{r=1}^{R} \mathbb{E}_{W_{r-1}}\Big[ \frac{1}{2} \log \Big\vert \frac{\eta_r^2}{\xi^2_r} \Sigma^r_t + I_d \Big\vert\Big] \nonumber\\
		\leq & \sum_{r=1}^{R}  \frac{1}{2} \log \Big\vert \frac{\eta_r^2}{\xi^2_r} \mathbb{E}_{W_{r-1}}\big[ \Sigma^r_t \big]  + I_d \Big\vert. \label{ineq:14}
	\end{align}
	Applying analogous analysis to $I(W_R;M^i)$, for all $i\in[t-1]$, 
	\begin{align}
		I(W_R;M^i) = & I\Big(W_{R-1} + \eta_R \big(\sum_{i=1}^{t-1}\tilde{g}_i(W_R) + g_t(W_R) \big)+N_R ;M^i \Big) \nonumber\\
		= & I(W_{R-1} + \eta_R \tilde{g}_i(W_R) +N_R ;M^i ) \nonumber\\
		\leq & I(W_{R-1} , \eta_R \tilde{g}_i(W_R) +N_R ; M^i) \nonumber\\
		\leq & \sum_{r=1}^{R} I(\eta_r \tilde{g}_i(W_r) +N_r ; M^i| W_{r-1}). \label{ineq:15}
	\end{align}
	Since  $N_r$ is independent of $M^i$ and $\tilde{B}^r_i$, we get that 
	\begin{equation*}
		\mathrm{Cov}_{\tilde{B}^r_i, N_r} [\eta_r \tilde{g}_i(W_r) +N_r ] = \mathrm{Cov}_{\tilde{B}^r_i}[\eta_r \tilde{g}_i(W_r)] +  \mathrm{Cov}_{N_r} [N_r] =  \eta_r^2 \tilde{\Sigma}^r_i + \xi^2_r I_d.
	\end{equation*}
	By using Lemma \ref{lemmaA.9} with $\Sigma = \eta_r^2 \tilde{\Sigma}^r_i + \xi^2_r I_d$, we similarly obtain
	\begin{align*}
		& I^{W_{r-1}}(\eta_r \tilde{g}_i(W_r) +N_r ; M^i) \\
		\leq & H(\eta_r \tilde{g}_i(W_r) +N_r |W_{r-1}) - H(\eta_r \tilde{g}_i(W_r) +N_r |\tilde{B}^r_i,W_{r-1}) \\
		= &  H(\eta_r \tilde{g}_i(W_r) +N_r |W_{r-1}) - H(N_r) \\
		\leq & \frac{d}{2} \log(2\pi e) + \frac{1}{2} \log \Big\vert \eta_r^2 \tilde{\Sigma}^r_i + \xi^2_r I_d \Big\vert - \frac{d}{2} \log (2\pi e \xi^2_r) \\
		\leq & \frac{1}{2} \log \Big\vert \frac{\eta_r^2}{\xi^2_r} \tilde{\Sigma}^r_i + I_d \Big\vert.
	\end{align*}
	Putting the above inequality back into (\ref{ineq:15}) implies that 
	\begin{align}
		I(W_R;M^i)\leq  &\sum_{r=1}^{R} \mathbb{E}_{W_{r-1}}\Big[ I^{W_{r-1}}(\eta_r \tilde{g}_i(W_r) +N_r ; M^i)\Big] \nonumber\\
		\leq & \sum_{r=1}^{R}  \frac{1}{2} \log \Big\vert \frac{\eta_r^2}{\xi^2_r} \mathbb{E}_{W_{r-1}}\big[ \tilde{\Sigma}^r_i \big]  + I_d \Big\vert. \label{ineq:16}
	\end{align}
	Substituting inequalities (\ref{ineq:14}) and (\ref{ineq:16}) into Theorem \ref{input-output},
	\begin{align*}
		\vert \mathrm{gen}_{W} \vert \leq &  \sqrt{\frac{2\sigma^2}{n} I(W;D^t)} + \sum_{i=1}^{t-1} \bigg(\sqrt{\frac{2\sigma^2}{n} I(W;M^i)} +\sqrt{\frac{2\sigma^2(n-\tilde{n})}{n\tilde{n}} I(W;M^i)}\bigg) \\
		\leq & \sqrt{\frac{\sigma^2}{n} \sum_{r=1}^{R} \log \Big\vert \frac{\eta_r^2}{\xi^2_r} \mathbb{E}_{W_{r-1}}\big[ \Sigma^r_t \big]  + I_d \Big\vert} + \sum_{i=1}^{t-1} \Bigg(\sqrt{\frac{\sigma^2}{n} \sum_{r=1}^{R}\log \Big\vert \frac{\eta_r^2}{\xi^2_r} \mathbb{E}_{W_{r-1}}\big[ \tilde{\Sigma}^r_i \big]  + I_d \Big\vert} \\
		& + \sqrt{\frac{\sigma^2(n-\tilde{n})}{n\tilde{n}} \sum_{r=1}^{R}\log \Big\vert \frac{\eta_r^2}{\xi^2_r} \mathbb{E}_{W_{r-1}}\big[ \tilde{\Sigma}^r_i \big]  + I_d \Big\vert} \Bigg) \\
		\leq & \sqrt{\frac{\sigma^2}{n} \sum_{r=1}^{R} \log \Big\vert \frac{\eta_r^2}{\xi^2_r} \mathbb{E}_{W_{r-1}}\big[ \Sigma^r_t \big]  + I_d \Big\vert} + \sum_{i=1}^{t-1} \sqrt{\frac{2\sigma^2}{\tilde{n}} \sum_{r=1}^{R}\log \Big\vert \frac{\eta_r^2}{\xi^2_r} \mathbb{E}_{W_{r-1}}\big[ \tilde{\Sigma}^r_i \big]  + I_d \Big\vert},
	\end{align*}
	where the last inequality is due to the triangle inequality. This completes the proof.
\end{proof}

\section{Detailed Proofs in Section \ref{Section4}}

\begin{restatetheorem}{\ref{e-cmi}}[Restate]
	Let $n$ and $\tilde{n}$ denote the number of samples available for training the current task and the number of the memory, respectively. Assume that $\ell(w,Z)$, where $Z\in\mathcal{Z}$, is $\sigma$-subgaussian for all $w\in\mathcal{W}$, we have 
	\begin{equation*}
		\vert \mathrm{gen}_{W} \vert \leq \underbrace{\mathbb{E}_{\tilde{D}^{t}} \sqrt{\frac{4\sigma^2 I^{\tilde{D}^t}(L^{i}; S^i)}{n}}}_{\text{Current Task Generalization}} + \underbrace{\mathbb{E}_{\tilde{D}^{1:t-1}} \sum_{i=1}^{t-1} \sqrt{\frac{4\sigma^2I^{\tilde{D}^i}(L^{i}; S^i)}{n}}}_{\text{Previous Task Generalization}} + \underbrace{\mathbb{E}_{\tilde{D}^{1:t-1},S^{1:t-1}} \sum_{i=1}^{t-1}\sqrt{\frac{2\sigma^2 (n-\tilde{n})}{n\tilde{n}} I^{\tilde{D}^{i}_{S^i}}(L^{i}; U^{i})}}_{\text{Memory Compression Cost}}.
	\end{equation*}
\end{restatetheorem}
\begin{proof}
	Notice that 
	\begin{align}
		\vert \mathrm{gen}_{W} \vert 
		= & \Bigg\vert \mathbb{E}_{W,\tilde{D}^{1:t},S^{1:t}, U^{1:t-1}}\Bigg[\sum_{i=1}^t \frac{1}{n} \sum_{j=1}^{n} \ell(W,\tilde{Z}^i_{j,1-S^i_j}) - \bigg(\sum_{i=1}^{t-1} \frac{1}{\tilde{n}}\sum_{j=1}^{n} U^i_j \ell(W,\tilde{Z}^i_{j,S^i_j}) + \frac{1}{n}\sum_{i=1}^n \ell(W,\tilde{Z}^i_{j,S^i_j})\bigg)  \nonumber\\
		& + \sum_{i=1}^{t-1} \frac{1}{n} \sum_{j=1}^{n} \ell(W,\tilde{Z}^i_{j,S^i_j}) -  \sum_{i=1}^{t-1} \frac{1}{n} \sum_{j=1}^{n} \ell(W,\tilde{Z}^i_{j,S^i_j}) \Bigg]\Bigg\vert\nonumber\\
		\leq & \mathbb{E}_{\tilde{D}^{1:t}} \Bigg\vert \mathbb{E}_{W,S^{1:t}|\tilde{D}^{1:t}}\Bigg[ \sum_{i=1}^t  \frac{1}{n} \sum_{j=1}^{n} \Big(\ell(W,\tilde{Z}^i_{j,1-S^i_j}) - \ell(W,\tilde{Z}^i_{j,S^i_j})\Big) \Bigg] \Bigg\vert \nonumber \\
		&  + \mathbb{E}_{\tilde{D}^{1:t-1},S^{1:t-1}} \Bigg\vert \mathbb{E}_{W, U^{1:t-1}|\tilde{D}^{1:t-1},S^{1:t-1}}\Bigg[\sum_{i=1}^{t-1} \Big( \frac{1}{n} \sum_{j=1}^{n} \ell(W,\tilde{Z}^i_{j,1-S^i_j}) -  \frac{1}{\tilde{n}}\sum_{j=1}^{n} U^i_j \ell(W,\tilde{Z}^i_{j,S^i_j})\Big) \Bigg] \Bigg\vert \nonumber \\
		\leq & \mathbb{E}_{\tilde{D}^{1:t}}  \sum_{i=1}^t  \Bigg\vert \mathbb{E}_{L^i,S^{i}|\tilde{D}^{i}}\Bigg[ \frac{1}{n} \sum_{j=1}^{n} \Big(\ell(W,\tilde{Z}^i_{j,1-S^i_j}) - \ell(W,\tilde{Z}^i_{j,S^i_j})\Big) \Bigg] \Bigg\vert \nonumber \\
		& + \mathbb{E}_{\tilde{D}^{1:t-1},S^{1:t-1}} \sum_{i=1}^{t-1}  \Bigg\vert \mathbb{E}_{L^i, U^{i}|\tilde{D}^{i},S^{i}}\Bigg[ \frac{1}{n} \sum_{j=1}^{n} \ell(W,\tilde{Z}^i_{j,1-S^i_j}) -  \frac{1}{\tilde{n}}\sum_{j=1}^{n} U^i_j \ell(W,\tilde{Z}^i_{j,S^i_j})\Bigg] \Bigg\vert. \label{ineq:17}
	\end{align}    
	Let $f(L^i,S^i) =  \frac{1}{n} \sum_{j=1}^{n} \big(\ell(W,\tilde{Z}^i_{j,1-S^i_j}) - \ell(W,\tilde{Z}^i_{j,S^i_j})\big) $. Due to the symmetry in the definition of the training and test sets, we have $\mathbb{E}_{\bar{S}^i}[f(L^i,\bar{S}^i)] = 0$. Furthermore, it is proven that $f(L^i,S^i)$ is a $\sigma\sqrt{\frac{2}{n}}$-subGaussian random variable. By leveraging Lemma \ref{lemmaA.6}, we get that 
	\begin{equation*}
		\Bigg\vert \mathbb{E}_{L^i,S^{i}|\tilde{D}^{i}}\Bigg[ \frac{1}{n} \sum_{j=1}^{n} \Big(\ell(W,\tilde{Z}^i_{j,1-S^i_j}) - \ell(W,\tilde{Z}^i_{j,S^i_j})\Big) \Bigg] \Bigg\vert \leq \sqrt{\frac{4\sigma^2}{n} I^{\tilde{D}^{i}}(L^i; S^{i})}.
	\end{equation*}
	Plugging the above inequality into the first term on the RHS of (\ref{ineq:17}), 
	\begin{equation}\label{ineq:18}
		\mathbb{E}_{\tilde{D}^{1:t}}  \sum_{i=1}^t  \Bigg\vert \mathbb{E}_{L^i,S^{i}|\tilde{D}^{i}}\Bigg[ \frac{1}{n} \sum_{j=1}^{n} \Big(\ell(W,\tilde{Z}^i_{j,1-S^i_j}) - \ell(W,\tilde{Z}^i_{j,S^i_j})\Big) \Bigg] \Bigg\vert \leq \mathbb{E}_{\tilde{D}^{1:t}}  \sum_{i=1}^t  \sqrt{\frac{4\sigma^2}{n} I^{\tilde{D}^{i}}(L^i; S^{i})}.
	\end{equation}
	Similarly, let $f(L^i,U^i) =  \frac{1}{n} \sum_{j=1}^{n} \ell(W,\tilde{Z}^i_{j,1-S^i_j}) -  \frac{1}{\tilde{n}}\sum_{j=1}^{n} U^i_j \ell(W,\tilde{Z}^i_{j,S^i_j})$. It can be shown that $\mathbb{E}_{\bar{U}^i}[f(L^i,\bar{U}^i)]=0$ and $f(L^i,U^i)$ is a $\sigma\sqrt{\frac{1}{\tilde{n}}-\frac{1}{n}}$-subGaussian random variable. Applying Lemma \ref{lemmaA.6} implies that 
	\begin{equation*}
		\Bigg\vert \mathbb{E}_{L^i, U^{i}|\tilde{D}^{i},S^{i}}\Bigg[ \frac{1}{n} \sum_{j=1}^{n} \ell(W,\tilde{Z}^i_{j,1-S^i_j}) -  \frac{1}{\tilde{n}}\sum_{j=1}^{n} U^i_j \ell(W,\tilde{Z}^i_{j,S^i_j})\Bigg] \Bigg\vert\leq \sqrt{\frac{2\sigma^2 (n-\tilde{n})}{n\tilde{n}} I^{\tilde{D}^{i}_{S^{i}}}(L^i; U^{i})}.
	\end{equation*}
	Putting this inequality back into the second term on the RHS of (\ref{ineq:17}), we have
	\begin{align}
		&\mathbb{E}_{\tilde{D}^{1:t-1},S^{1:t-1}} \sum_{i=1}^{t-1}  \Bigg\vert \mathbb{E}_{L^i, U^{i}|\tilde{D}^{i},S^{i}}\Bigg[ \frac{1}{n} \sum_{j=1}^{n} \ell(W,\tilde{Z}^i_{j,1-S^i_j}) -  \frac{1}{\tilde{n}}\sum_{j=1}^{n} U^i_j \ell(W,\tilde{Z}^i_{j,S^i_j})\Bigg] \Bigg\vert \nonumber\\
		\leq & \mathbb{E}_{\tilde{D}^{1:t-1},S^{1:t-1}} \sum_{i=1}^{t-1}  \sqrt{\frac{2\sigma^2 (n-\tilde{n})}{n\tilde{n}} I^{\tilde{D}^{i}_{S^{i}}}(L^i; U^{i})}. \label{ineq:19}
	\end{align}
	Combining inequalities (\ref{ineq:18}), (\ref{ineq:19}), and (\ref{ineq:17}), we obtain that 
	\begin{equation*}
		\vert \mathrm{gen}_{W} \vert \leq \mathbb{E}_{\tilde{D}^{t}} \sqrt{\frac{4\sigma^2 I^{\tilde{D}^t}(L^i; S^i)}{n}} + \mathbb{E}_{\tilde{D}^{1:t-1}} \sum_{i=1}^{t-1} \sqrt{\frac{4\sigma^2I^{\tilde{D}^i}(L^i; S^i)}{n}} + \mathbb{E}_{\tilde{D}^{1:t-1},S^{1:t-1}} \sum_{i=1}^{t-1}\sqrt{\frac{2\sigma^2 (n-\tilde{n})}{n\tilde{n}} I^{\tilde{D}^{i}_{S^i}}(L^i; U^{i})},
	\end{equation*}
\end{proof}

\subsection{Proof of Theorem \ref{e-mi}}

\begin{restatetheorem}{\ref{e-mi}}[Restate]
	Let $n$ and $\tilde{n}$ denote the number of samples available for training the current task and the number of the memory, respectively. Assume that $\ell(w,Z)$, where $Z\in\mathcal{Z}$, is $\sigma$-subgaussian for all $w\in\mathcal{W}$, we have 
	\begin{equation*}
		\vert \mathrm{gen}_{W} \vert \leq \underbrace{\sqrt{\frac{4\sigma^2 I(L^i; S^{i})}{n}}}_{\text{Current Task Generalization}} + \sum_{i=1}^{t-1} \Bigg( \underbrace{ \sqrt{\frac{4\sigma^2I(L^i; S^i)}{n}}}_{\text{Previous Task Generalization}} + \underbrace{ \sqrt{\frac{2\sigma^2 (n-\tilde{n})I(L^{i,S^i}; U^{i})}{n\tilde{n}} }}_{\text{Memory Compression Cost}} \Bigg).
	\end{equation*}
\end{restatetheorem}
\begin{proof}
	Notably, if we do not move the expectation over $\tilde{D}^{1:t-1}$ (or $\tilde{D}^{1:t-1}, S^{1:t-1}$) outside of the absolute function, we have a chance to get ride of the expectation over $\tilde{D}^{1:t-1}$ (or $\tilde{D}^{1:t-1}, S^{1:t-1}$) if we take the expectation over $L^i$ (or $L^{i,S^i}$). By the definition of the generalization error, we have 
	\begin{align}
		\vert \mathrm{gen}_{W} \vert 
		= & \Bigg\vert \mathbb{E}_{W,\tilde{D}^{1:t},S^{1:t}, U^{1:t-1}}\Bigg[\sum_{i=1}^t \frac{1}{n} \sum_{j=1}^{n} \ell(W,\tilde{Z}^i_{j,1-S^i_j}) - \bigg(\sum_{i=1}^{t-1} \frac{1}{\tilde{n}}\sum_{j=1}^{n} U^i_j \ell(W,\tilde{Z}^i_{j,S^i_j}) + \frac{1}{n}\sum_{i=1}^n \ell(W,\tilde{Z}^i_{j,S^i_j})\bigg)  \nonumber\\
		& + \sum_{i=1}^{t-1} \frac{1}{n} \sum_{j=1}^{n} \ell(W,\tilde{Z}^i_{j,S^i_j}) -  \sum_{i=1}^{t-1} \frac{1}{n} \sum_{j=1}^{n} \ell(W,\tilde{Z}^i_{j,S^i_j}) \Bigg]\Bigg\vert\nonumber\\
		\leq &  \Bigg\vert \mathbb{E}_{W,\tilde{D}^{1:t},S^{1:t}}\Bigg[ \sum_{i=1}^t  \frac{1}{n} \sum_{j=1}^{n} \Big(\ell(W,\tilde{Z}^i_{j,1-S^i_j}) - \ell(W,\tilde{Z}^i_{j,S^i_j})\Big) \Bigg] \Bigg\vert \nonumber \\
		&  +  \Bigg\vert \mathbb{E}_{W,\tilde{D}^{1:t-1},S^{1:t-1},U^{1:t-1}}\Bigg[\sum_{i=1}^{t-1} \Big( \frac{1}{n} \sum_{j=1}^{n} \ell(W,\tilde{Z}^i_{j,1-S^i_j}) -  \frac{1}{\tilde{n}}\sum_{j=1}^{n} U^i_j \ell(W,\tilde{Z}^i_{j,S^i_j})\Big) \Bigg] \Bigg\vert \nonumber \\
		\leq &  \sum_{i=1}^t  \Bigg\vert \mathbb{E}_{L^i,S^{i}}\Bigg[ \frac{1}{n} \sum_{j=1}^{n} \Big(\ell(W,\tilde{Z}^i_{j,1-S^i_j}) - \ell(W,\tilde{Z}^i_{j,S^i_j})\Big) \Bigg] \Bigg\vert \nonumber \\
		& + \sum_{i=1}^{t-1}  \Bigg\vert \mathbb{E}_{L^{i,S^i}, U^{i}}\Bigg[ \frac{1}{n} \sum_{j=1}^{n} \ell(W,\tilde{Z}^i_{j,1-S^i_j}) -  \frac{1}{\tilde{n}}\sum_{j=1}^{n} U^i_j \ell(W,\tilde{Z}^i_{j,S^i_j})\Bigg] \Bigg\vert. \label{ineq:20}
	\end{align}   
	Applying the analogous analysis of Theorem \ref{e-cmi} and using Lemma \ref{lemmaA.6} with $f(L^i,S^{i}) = \frac{1}{n} \sum_{j=1}^{n} \big(\ell(W,\tilde{Z}^i_{j,1-S^i_j}) - \ell(W,\tilde{Z}^i_{j,S^i_j})\big)$, we get that 
	\begin{equation}\label{ineq:21}
		\Bigg\vert \mathbb{E}_{L^i,S^{i}}\Bigg[ \frac{1}{n} \sum_{j=1}^{n} \Big(\ell(W,\tilde{Z}^i_{j,1-S^i_j}) - \ell(W,\tilde{Z}^i_{j,S^i_j})\Big) \Bigg] \Bigg\vert \leq \sqrt{\frac{4\sigma^2}{n} I(L^i; S^{i})},
	\end{equation}
	since $f(L^i,S^{i})$ is $\sigma\sqrt{\frac{2}{n}}$-subGaussian and $\mathbb{E}_{\bar{S}^i}[f(L^i,\bar{S}^i)]=0$. Similarly, by leveraging Lemma \ref{lemmaA.6} with $f(L^{i,S^i},U^i) =  \frac{1}{n} \sum_{j=1}^{n} \ell(W,\tilde{Z}^i_{j,1-S^i_j}) -  \frac{1}{\tilde{n}}\sum_{j=1}^{n} U^i_j \ell(W,\tilde{Z}^i_{j,S^i_j})$, we have 
	\begin{equation}\label{ineq:22}
		\Bigg\vert \mathbb{E}_{L^{i,S^i}, U^{i}}\Bigg[ \frac{1}{n} \sum_{j=1}^{n} \ell(W,\tilde{Z}^i_{j,1-S^i_j}) -  \frac{1}{\tilde{n}}\sum_{j=1}^{n} U^i_j \ell(W,\tilde{Z}^i_{j,S^i_j})\Bigg] \Bigg\vert\leq \sqrt{\frac{2\sigma^2 (n-\tilde{n})}{n\tilde{n}} I(L^{i,S^i}; U^{i})}.
	\end{equation}
	Substituting inequalities (\ref{ineq:21}) and (\ref{ineq:21}) into (\ref{ineq:20}), we obtain
	\begin{equation*}
		\vert \mathrm{gen}_{W} \vert \leq  \sqrt{\frac{4\sigma^2 I(L^i; S^{i})}{n}} +  \sum_{i=1}^{t-1} \Bigg( \sqrt{\frac{4\sigma^2I(L^i; S^i)}{n}} +  \sqrt{\frac{2\sigma^2 (n-\tilde{n})I(L^{i,S^i}; U^{i})}{n\tilde{n}} }\Bigg),
	\end{equation*}
	which completes the proof.
\end{proof}

\subsection{Proof of Theorem \ref{binarykl2}}

\begin{restatetheorem}{\ref{binarykl2}}[Restate]
	Let $n$ and $\tilde{n}$ denote the number of samples available for training the current task and the number of the memory, respectively. Assume that $\ell(\cdot,\cdot) \in [0,1]$, we have
	\begin{equation*}
		d\bigg(\hat{R} \bigg\Vert \frac{\hat{R}+R}{2} \bigg)  \leq \sum_{i=1}^{t-1}  \frac{1}{\tilde{n}}\sum_{j:U^i_j=1}I(L^i_j;S^i_j|U^i_j=1)  +  \frac{1}{n}\sum_{j=1}^{n} I(L^t_j;S^t_j).
	\end{equation*}
\end{restatetheorem}

\begin{proof}
	We turn to immediately bounding the expected generalization error. By Jensen's inequality and the joint convexity of $d_\gamma(\cdot)$, we have 
	\begin{align}
		&d\bigg(\hat{R} \bigg\Vert \frac{\hat{R}+R}{2} \bigg) \nonumber\\
		\leq &  \sup_\gamma d_\gamma\Bigg(\mathbb{E}_{W,\tilde{D}^{1:t-1},S^{1:t-1}, U^{1:t-1}}\sum_{i=1}^{t-1} \frac{1}{\tilde{n}}\sum_{j:U^i_j=1}\ell(W,\tilde{Z}^i_{j,S^i_j})+ \mathbb{E}_{W,\tilde{D}^t,S^t} \frac{1}{n}\sum_{j=1}^{n} \ell(W,\tilde{Z}^t_{j,S^t_j}) \bigg\Vert  \nonumber\\
		& \mathbb{E}_{W,\tilde{D}^{1:t-1},S^{1:t-1}, U^{1:t-1}}\sum_{i=1}^{t-1} \frac{1}{\tilde{n}}\sum_{j:U^i_j=1} \frac{\ell(W,\tilde{Z}^i_{j,1})+ \ell(W,\tilde{Z}^i_{j,0})}{2} +   \mathbb{E}_{W,\tilde{D}^t,S^t} \frac{1}{n}\sum_{j=1}^{n} \frac{\ell(W,\tilde{Z}^t_{j,1})+ \ell(W,\tilde{Z}^t_{j,0})}{2}   \Bigg) \nonumber\\
		\leq & \sup_\gamma \Bigg\{ \mathbb{E}_{U^{1:t-1}} \sum_{i=1}^{t-1}  \frac{1}{\tilde{n}}\sum_{j:U^i_j=1} d_\gamma \Bigg( \mathbb{E}_{W, \tilde{Z}^i_j,S^i_j| U^i_j=1} \big[\ell(W,\tilde{Z}^i_{j,S^i_j})\big] \bigg\Vert \mathbb{E}_{W, \tilde{Z}^i_j,S^i_j| U^i_j=1} \bigg[\frac{\ell(W,\tilde{Z}^i_{j,1})+ \ell(W,\tilde{Z}^i_{j,0})}{2}\bigg]  \Bigg)  \nonumber\\
		& +  \frac{1}{n}\sum_{j=1}^{n} d_\gamma \Bigg( \mathbb{E}_{W, \tilde{Z}^t_j,S^t_j} \big[\ell(W,\tilde{Z}^t_{j,S^t_j})\big] \bigg\Vert \mathbb{E}_{W, \tilde{Z}^t_j,S^t_j} \bigg[\frac{\ell(W,\tilde{Z}^t_{j,1})+ \ell(W,\tilde{Z}^t_{j,0})}{2}\bigg]  \Bigg) \Bigg\} \nonumber\\
		= & \sup_\gamma \Bigg\{ \mathbb{E}_{U^{1:t-1}} \sum_{i=1}^{t-1}  \frac{1}{\tilde{n}}\sum_{j:U^i_j=1} d_\gamma \Bigg( \mathbb{E}_{L^i_j,S^i_j| U^i_j=1} \big[\ell(W,\tilde{Z}^i_{j,S^i_j})\big] \bigg\Vert \mathbb{E}_{L^{i}_j,S^i_j| U^i_j=1} \bigg[\frac{\ell(W,\tilde{Z}^i_{j,1})+ \ell(W,\tilde{Z}^i_{j,0})}{2}\bigg]  \Bigg)   \nonumber\\
		& +  \frac{1}{n}\sum_{j=1}^{n} d_\gamma \Bigg( \mathbb{E}_{L^t_j,S^t_j} \big[\ell(W,\tilde{Z}^t_{j,S^t_j})\big] \bigg\Vert \mathbb{E}_{L^t_j,S^t_j} \bigg[\frac{\ell(W,\tilde{Z}^t_{j,1})+ \ell(W,\tilde{Z}^t_{j,0})}{2}\bigg]  \Bigg) \Bigg\} \nonumber\\
		\leq & \sup_\gamma \bigg\{ \mathbb{E}_{U^{1:t-1}} \sum_{i=1}^{t-1}  \frac{1}{\tilde{n}}\sum_{j:U^i_j=1} \mathbb{E}_{L^i_j,S^i_j| U^i_j=1} \bigg[d_\gamma\bigg(L^i_{j,S^i_j}\bigg\Vert \frac{L^i_{j,0}+ L^i_{j,1}}{2}\bigg)\bigg]   +  \frac{1}{n}\sum_{j=1}^{n} \mathbb{E}_{L^t_j,S^t_j} \bigg[d_\gamma\bigg(L^t_{j,S^t_j}\bigg\Vert \frac{L^t_{j,0}+ L^t_{j,1}}{2}\bigg)\bigg]  \bigg\} \label{ineq:23}
	\end{align}
	For all $i\in[t-1]$ and any $\gamma>0$, by employing Lemma \ref{lemmaA.5} with $P=P_{L^i_j,S^i_j| U^i_j=1}$, $Q=P_{L^i_j|U^i_j=1} P_{S^i_j} $, and $f(L^i_j,S^i_j) = d_\gamma\Big(L^i_{j,S^i_j}\Big\Vert \frac{L^i_{j,0}+ L^i_{j,1}}{2}\Big)$, we have 
	\begin{align}
		I^{U^i_j=1}(L^i_j;S^i_j) = &  D(P_{L^i_j,S^i_j| U^i_j=1} \Vert P_{L^i_j|U^i_j=1} P_{S^i_j}) \nonumber\\
		\geq & \mathbb{E}_{L^i_j,S^i_j| U^i_j=1} \Big [d_\gamma\Big(L^i_{j,S^i_j}\Big\Vert \frac{L^i_{j,0}+ L^i_{j,1}}{2}\Big) \Big] - \log \mathbb{E}_{L^i_j,\bar{S}^i_j| U^i_j=1} \Big[e^{d_\gamma\Big(L^i_{j,\bar{S}^i_j}\Big\Vert \frac{L^i_{j,0}+ L^i_{j,1}}{2}\Big) }\Big].\label{ineq:24}
	\end{align}
	Note that $\mathbb{E}_{L^i_j,\bar{S}^i_j| U^i_j=1}[L^i_{j,\bar{S}^i_j}] = \frac{L^i_{j,0}+ L^i_{j,1}}{2}$. Applying Lemma \ref{lemmaA.8}, for any  $\gamma>0$, we have 
	\begin{equation*}
		\mathbb{E}_{L^i_j,\bar{S}^i_j| U^i_j=1} \Big[e^{d_\gamma\Big(L^i_{j,\bar{S}^i_j}\Big\Vert \frac{L^i_{j,0}+ L^i_{j,1}}{2}\Big) }\Big] \leq 1.
	\end{equation*}
	Putting the above inequality back into (\ref{ineq:24}), we obtain
	\begin{equation}\label{ineq:25}
		\mathbb{E}_{L^i_j,S^i_j| U^i_j=1} \Big [d_\gamma\Big(L^i_{j,S^i_j}\Big\Vert \frac{L^i_{j,0}+ L^i_{j,1}}{2}\Big) \Big] \leq I^{U^i_j=1}(L^i_j;S^i_j).
	\end{equation}
	Furthermore, for any $\gamma>0$, we utilize Lemma \ref{lemmaA.5} with $P=P_{L^t_j,S^t_j}$, $Q=P_{L^t_j} P_{S^t_j} $, and $f(L^t_j,S^t_j) = d_\gamma\Big(L^t_{j,S^t_j}\Big\Vert \frac{L^t_{j,0}+ L^t_{j,1}}{2}\Big)$, having
	\begin{equation}\label{ineq:26}
		I(L^t_j;S^t_j) = D(P_{L^t_j,S^t_j} \Big\Vert P_{L^t_j} P_{S^t_j})  \geq  \mathbb{E}_{L^t_j,S^t_j} \Big [d_\gamma\Big(L^t_{j,S^t_j}\Big\Vert \frac{L^t_{j,0}+ L^t_{j,1}}{2}\Big) \Big] - \log \mathbb{E}_{L^t_j,\bar{S}^t_j} \Big[e^{d_\gamma\Big(L^t_{j,\bar{S}^t_j}\Big\Vert \frac{L^t_{j,0}+ L^t_{j,1}}{2}\Big) }\Big].
	\end{equation}
	Similarly, we prove that $\mathbb{E}_{L^t_j,\bar{S}^t_j} \big[e^{d_\gamma\big(L^t_{j,\bar{S}^t_j}\big\Vert \frac{L^t_{j,0}+ L^t_{j,1}}{2}\big) }\big] \leq 1$ and plugging it into the inequality (\ref{ineq:26}), implying 
	\begin{equation}\label{ineq:27}
		\mathbb{E}_{L^t_j,S^t_j} \Big [d_\gamma\Big(L^t_{j,S^t_j}\Big\Vert \frac{L^t_{j,0}+ L^t_{j,1}}{2}\Big) \Big]\leq I(L^t_j;S^t_j).
	\end{equation}
	Substituting inequalities (\ref{ineq:27}) and (\ref{ineq:25}) into (\ref{ineq:23}), we obtain that 
	\begin{equation}
		d\bigg(\hat{R} \bigg\Vert \frac{\hat{R}+R}{2} \bigg)  \leq \sum_{i=1}^{t-1}  \frac{1}{\tilde{n}}\sum_{j:U^i_j=1}I(L^i_j;S^i_j|U^i_j=1)  +  \frac{1}{n}\sum_{j=1}^{n} I(L^t_j;S^t_j).
	\end{equation}
\end{proof}

\subsection{Proof of Theorem \ref{ld-mi}}

\begin{restatetheorem}{\ref{ld-mi}}[Restate]
	Let $n$ and $\tilde{n}$ denote the number of samples available for training the current task and the number of the memory, respectively. Assume that $\ell(\cdot,\cdot) \in [0,1]$, we have
	\begin{equation*}
		\vert \mathrm{gen}_W\vert \leq  \mathbb{E}_{U^{1:t-1}} \sum_{i=1}^{t-1} \frac{1}{\tilde{n}}\sum_{j:U^i_j=1} \sqrt{2I^{U^i_j=1}(\Delta^i_j;S^i_j)} + \frac{1}{n}\sum_{j=1}^{n} \sqrt{2I(\Delta^t_j;S^t_j)}.
	\end{equation*}
\end{restatetheorem}
\begin{proof}
	By the definition of the generalization error and Jensen's inequality,
	\begin{align}
		\vert \mathrm{gen}_W\vert 
		\leq & \mathbb{E}_{U^{1:t-1}}\sum_{i=1}^{t-1} \frac{1}{\tilde{n}}\sum_{j:U^i_j=1} \bigg\vert\mathbb{E}_{W,\tilde{Z}^i_j,S^i_j| U^i_j=1} \Big[\ell(W,\tilde{Z}^i_{j,{1-S^i_j}}) - \ell(W,\tilde{Z}^i_{j,S^i_j}) \Big] \bigg\vert \nonumber\\
		& + \frac{1}{n}\sum_{j=1}^{n} \bigg\vert\mathbb{E}_{W,\tilde{Z}^i_j,S^i_j} \Big[\ell(W,\tilde{Z}^t_{j,{1-S^t_j}}) - \ell(W,\tilde{Z}^t_{j,S^t_j}) \Big]\bigg\vert \nonumber\\
		= & \mathbb{E}_{U^{1:t-1}}\sum_{i=1}^{t-1} \frac{1}{\tilde{n}}\sum_{j:U^i_j=1} \bigg\vert\mathbb{E}_{L^i_j,S^i_j| U^i_j=1} \Big[L^i_{j,\bar{S}^i_j} - L^i_{j,S^i_j} \Big] \bigg\vert  + \frac{1}{n}\sum_{j=1}^{n} \bigg\vert\mathbb{E}_{L^t_j,S^t_j} \Big[L^t_{j,\bar{S}^t_j} - L^t_{j,S^t_j} \Big]\bigg\vert \nonumber\\
		= & \mathbb{E}_{U^{1:t-1}}\sum_{i=1}^{t-1} \frac{1}{\tilde{n}}\sum_{j:U^i_j=1} \bigg\vert\mathbb{E}_{L^i_j,S^i_j| U^i_j=1} \Big[(-1)^{S^i_j}(L^i_{j,1} - L^i_{j,0}) \Big] \bigg\vert  + \frac{1}{n}\sum_{j=1}^{n} \bigg\vert\mathbb{E}_{L^t_j,S^t_j} \Big[(-1)^{S^t_j}(L^t_{j,1} - L^t_{j,0}) \Big]\bigg\vert \nonumber\\
		\leq & \mathbb{E}_{U^{1:t-1}}\sum_{i=1}^{t-1} \frac{1}{\tilde{n}}\sum_{j:U^i_j=1} \bigg\vert\mathbb{E}_{\Delta^i_j,S^i_j| U^i_j=1} \Big[(-1)^{S^i_j}\Delta^i_j \Big] \bigg\vert  + \frac{1}{n}\sum_{j=1}^{n} \bigg\vert\mathbb{E}_{\Delta^t_j,S^t_j} \Big[(-1)^{S^t_j}\Delta^t_j \Big]\bigg\vert \label{ineq:29}
	\end{align}
	For all $i\in[t-1]$, let $f(\Delta^i_j,S^i_j)=(-1)^{S^i_j}\Delta^i_j$. Since $\Delta^i_j\in[-1,+1]$, $f(\Delta^i_j,S^i_j)$ is a $1$-subGaussian random variable. Furthermore, $\mathbb{E}_{\bar{S}^i_j}f(\Delta^i_j,\bar{S}^i_j) = 0$.  We employ Lemma \ref{lemmaA.6} with $f(\Delta^i_j,S^i_j)$, having
	\begin{equation} \label{ineq:30}
		\bigg\vert\mathbb{E}_{\Delta^i_j,S^i_j| U^i_j=1} \Big[(-1)^{S^i_j}\Delta^i_j \Big] - \mathbb{E}_{\Delta^i_j,\bar{S}^i_j| U^i_j=1} \Big[(-1)^{\bar{S}^i_j}\Delta^i_j \Big]  \bigg\vert = \bigg\vert\mathbb{E}_{\Delta^i_j,S^i_j| U^i_j=1} \Big[(-1)^{S^i_j}\Delta^i_j \Big] \bigg\vert \leq \sqrt{2I^{U^i_j=1}(\Delta^i_j;S^i_j)}.
	\end{equation}
	Similarly, by leveraging Lemma \ref{lemmaA.6} with $f(\Delta^t_j,S^t_j)=(-1)^{S^t_j}\Delta^t_j$, we obtain
	\begin{equation}\label{ineq:31}
		\bigg\vert\mathbb{E}_{\Delta^t_j,S^t_j} \Big[(-1)^{S^t_j}\Delta^t_j \Big] - \mathbb{E}_{\Delta^t_j,\bar{S}^t_j} \Big[(-1)^{\bar{S}^t_j}\Delta^t_j \Big]  \bigg\vert = \bigg\vert\mathbb{E}_{\Delta^t_j,S^t_j} \Big[(-1)^{S^t_j}\Delta^t_j \Big] \bigg\vert  \leq \sqrt{2I(\Delta^t_j;S^t_j)}.
	\end{equation}
	Substituting inequalities (\ref{ineq:30}) and (\ref{ineq:31}) into (\ref{ineq:29}), we have 
	\begin{align*}
		\vert \mathrm{gen}_W\vert \leq & \mathbb{E}_{U^{1:t-1}}\sum_{i=1}^{t-1} \frac{1}{\tilde{n}}\sum_{j:U^i_j=1} \bigg\vert\mathbb{E}_{\Delta^i_j,S^i_j| U^i_j=1} \Big[(-1)^{S^i_j}\Delta^i_j \Big] \bigg\vert  + \frac{1}{n}\sum_{j=1}^{n} \bigg\vert\mathbb{E}_{\Delta^t_j,S^t_j} \Big[(-1)^{S^t_j}\Delta^t_j \Big]\bigg\vert \\
		\leq & \mathbb{E}_{U^{1:t-1}} \sum_{i=1}^{t-1} \frac{1}{\tilde{n}}\sum_{j:U^i_j=1} \sqrt{2I^{U^i_j=1}(\Delta^i_j;S^i_j)} + \frac{1}{n}\sum_{j=1}^{n} \sqrt{2I(\Delta^t_j;S^t_j)}.
	\end{align*}
	This completes the proof.

\end{proof}
\subsection{Proof of Theorem \ref{relia-fast}}
\begin{restatetheorem}{\ref{relia-fast}}[Restate]
	Let $n$ and $\tilde{n}$ denote the number of samples available for training the current task and the number of the memory, respectively. Assume that $\ell(\cdot,\cdot) \in \{0,1\}$. In the interpolating setting when $\hat{R}=0$, we have 
	\begin{align*}
		R = & \sum_{i=1}^{t-1} \frac{1}{\tilde{n}}\sum_{j:U^i_j=1} \frac{I(\Delta^i_j;S^i_j|U^i_j=1)}{\log 2} + \frac{1}{n}\sum_{j=1}^{n} \frac{I(\Delta^t_j;S^t_j)}{\log 2}
		=  \sum_{i=1}^{t-1} \frac{1}{\tilde{n}}\sum_{j:U^i_j=1} \frac{I(L^i_j;S^i_j|U^i_j=1)}{\log 2} + \frac{1}{n}\sum_{j=1}^{n} \frac{I(L^t_j;S^t_j)}{\log 2}  \\
		\leq & \sum_{i=1}^{t-1} \frac{1}{\tilde{n}}\sum_{j:U^i_j=1} \frac{2I(L^i_{j,1};S^i_j|U^i_j=1)}{\log 2} + \frac{1}{n}\sum_{j=1}^{n} \frac{2I(L^t_{j,1};S^t_j)}{\log 2}.
	\end{align*}
\end{restatetheorem}

\begin{proof}
	According to the assumption that $\ell(\cdot,\cdot)\in\{0,1\}$ and $\hat{R}=0$, we get 
	\begin{align}
		R =  &  \mathbb{E}_{W,\tilde{D}^{1:t-1},S^{1:t-1}, U^{1:t-1}}\Big[\sum_{i=1}^{t-1} \frac{1}{\tilde{n}}\sum_{j:U^i_j=1} \ell(W,\tilde{Z}^i_{j,{1-S^i_j}}) \Big]  + \mathbb{E}_{W,\tilde{D}^t,S^t} \Big[ \frac{1}{n}\sum_{j=1}^{n}  \ell(W,\tilde{Z}^t_{j,{1-S^t_j}}) \Big] \bigg\vert \nonumber\\
		= & \mathbb{E}_{U^{1:t-1}}\sum_{i=1}^{t-1} \frac{1}{\tilde{n}}\sum_{j:U^i_j=1} \mathbb{E}_{L^i_j,S^i_j|U^i_j=1}\Big[L^i_{j,\bar{S}^i_j}\Big] + \frac{1}{n}\sum_{j=1}^{n} \mathbb{E}_{L^t_j,S^t_j}\Big[L^t_{j,\bar{S}^t_j}\Big] \nonumber\\
		= &  \mathbb{E}_{U^{1:t-1}}\sum_{i=1}^{t-1} \frac{1}{\tilde{n}}\sum_{j:U^i_j=1} \frac{\mathbb{E}_{L^i_{j,1}|S^i_j=0, U^i_j=1}\big[L^i_{j,1}\big]+\mathbb{E}_{L^i_{j,0}|S^i_j=1, U^i_j=1}\big[L^i_{j,1}\big]}{2} + \frac{1}{n}\sum_{j=1}^{n} \frac{\mathbb{E}_{L^t_{j,1}|S^t_j=0}\big[L^t_{j,1}\big]+\mathbb{E}_{L^t_{j,0}|S^t_j=1}\big[L^i_{j,1}\big]}{2} \nonumber\\
		=& \mathbb{E}_{U^{1:t-1}}\sum_{i=1}^{t-1} \frac{1}{\tilde{n}}\sum_{j:U^i_j=1} \frac{P(\Delta^i_j = 1 |S^i_j=0, U^i_j=1) + P(\Delta^i_j = -1 |S^i_j=1, U^i_j=1)}{2} \nonumber \\
		&  + \frac{1}{n}\sum_{j=1}^{n} \frac{P(\Delta^t_j = 1 |S^t_j=0) + P(\Delta^t_j = -1 |S^t_j=1)}{2}. \label{ineq:32}
	\end{align}
	It is noteworthy that for all $i\in[t]$, the distribution of training loss $L^i_{j,S^i_j}$ (or test loss $L^i_{j,\bar{S}^i_j}$) should be identical regardless of the value of $S^i_j$. Given $S^i_j$ and $U^i_j=1$ (or $S^t_j$), the distributions of $L^i_{j,S^i_j}$ and $L^i_{j,\bar{S}^i_j}$ are symmetric, namely, $P(L^i_{j,1} |S^i_j=0,U^i_j=1) = P(L^i_{j,0} |S^i_j=1,U^i_j=1)$ and $P(L^i_{j,1} |S^i_j=1,U^i_j=1) = P(L^i_{j,0} |S^i_j=0,U^i_j=1)$ \big(or $P(L^t_{j,1} |S^t_j=0) = P(L^t_{j,0} |S^t_j=1)$ and $P(L^t_{j,1} |S^t_j=1) = P(L^t_{j,0} |S^t_j=0)$\big). Let $\alpha_{i,j}= P(\Delta^i_j = 1 |S^i_j=0,U^i_j=1)$, then $P(\Delta^i_j = 0 |S^i_j=0,U^i_j=1)=1-\alpha_{i,j}$ and 
	\begin{align*}
		I^{U^i_j=1}(\Delta^i_j;S^i_j) = & H(\Delta^i_j|U^i_j=1) - H(\Delta^i_j|S^i_j,U^i_j=1) \\
		= & H(\frac{\alpha_{i,j}}{2},1-\alpha_{i,j},\frac{\alpha_{i,j}}{2}) - H(\alpha_{i,j}, 1-\alpha_{i,j}) \\
		= & -\alpha_{i,j} \log(\frac{\alpha_{i,j}}{2}) + \alpha_{i,j} \log(\alpha_{i,j}) = \alpha_{i,j} \log 2.
	\end{align*}
	Plugging $P(\Delta^i_j = 1 |S^i_j=0,U^i_j=1)= \alpha_{i,j} = I^{U^i_j=1}(\Delta^i_j;S^i_j)/\log 2$ into the first term on the RHS of (\ref{ineq:32}), then 
	\begin{equation}\label{eq:33}
		\mathbb{E}_{U^{1:t-1}}\sum_{i=1}^{t-1} \frac{1}{\tilde{n}}\sum_{j:U^i_j=1} \frac{P(\Delta^i_j = 1 |S^i_j=0, U^i_j=1) + P(\Delta^i_j = -1 |S^i_j=1, U^i_j=1)}{2} = \sum_{i=1}^{t-1} \frac{1}{\tilde{n}}\sum_{j:U^i_j=1} \frac{I(\Delta^i_j;S^i_j|U^i_j=1)}{\log 2}. 
	\end{equation}
	Applying the analogous analysis of (\ref{eq:33}), we similarly obtain that 
	\begin{equation}\label{eq:34}
		\frac{1}{n}\sum_{j=1}^{n} \frac{P(\Delta^t_j = 1 |S^t_j=0) + P(\Delta^t_j = -1 |S^t_j=1)}{2} = \frac{1}{n}\sum_{j=1}^{n} \frac{I(\Delta^t_j;S^t_j)}{\log 2}. 
	\end{equation}
	Putting equations (\ref{eq:33}) and (\ref{eq:34}) back into (\ref{ineq:32}), we get 
	\begin{equation}
		R = \sum_{i=1}^{t-1} \frac{1}{\tilde{n}}\sum_{j:U^i_j=1} \frac{I(\Delta^i_j;S^i_j|U^i_j=1)}{\log 2} + \frac{1}{n}\sum_{j=1}^{n} \frac{I(\Delta^t_j;S^t_j)}{\log 2}.
	\end{equation}
	Notably, for all $i\in[t-1]$, since $\hat{R}=0$, we know that $P(L^i_{j,0}=1, L^i_{j,1}=1 |U^i_j=1)=0$, and then have a bijection between $\Delta^i_j$ and $L^i_j$ given $U^i_j=1$: $\Delta^i_j = 0 \leftrightarrow L^i_j= \{0,0\}$, $\Delta^i_j  = 1 \leftrightarrow L^i_j= \{1,0\}$, and $\Delta^i_j= -1 \leftrightarrow L^i_j= \{0,1\}$. By the data-processing inequality, we thus obtain that $I^{U^i_j=1}(\Delta^i_j;S^i_j) = I^{U^i_j=1}(L^i_j;S^i_j)$, and 
	\begin{align*}
		I^{U^i_j=1}(L^i_{j,1};S^i_j) = & H(L^i_{j,1}|U^i_j=1) - H(L^i_{j,1}|S^i_j,U^i_j=1)= H(\frac{\alpha_{i,j}}{2},1-\frac{\alpha_{i,j}}{2}) - \frac{1}{2}H(\alpha_{i,j},1-\alpha_{i,j})\\
		=  & -\frac{\alpha_{i,j}}{2} \log(\frac{\alpha_{i,j}}{2}) - \big( 1- \frac{\alpha_{i,j}}{2}\big) \log\big( 1- \frac{\alpha_{i,j}}{2}\big)  +  \frac{\alpha_{i,j}}{2} \log(\alpha_{i,j}) + \frac{1-\alpha_{i,j}}{2} \log\big( 1- \alpha_{i,j}\big) \\
		\geq & -\frac{\alpha_{i,j}}{2} \log(\frac{\alpha_{i,j}}{2}) + \frac{\alpha_{i,j}}{2} \log(\alpha_{i,j}) = \frac{\alpha_{i,j}}{2}  \log 2,
	\end{align*}
	where the last inequality is due to convex function $f(x)=(1-x)\log(1-x)$ and Jensen's inequality $\frac{f(0)+f(\alpha_{i,j})}{2}\geq \frac{f(\alpha_{i,j})}{2}$. Combining the above estimates, we get that 
	\begin{equation}\label{ineq:36}
		I^{U^i_j=1}(\Delta^i_j;S^i_j) = I^{U^i_j=1}(L^i_j;S^i_j) \leq  2  I^{U^i_j=1}(L^i_{j,1};S^i_j).
	\end{equation}
	Similarly, it can be proven that 
	\begin{equation}\label{ineq:37}
		I(\Delta^t_j;S^t_j) = I(L^t_j;S^t_j) \leq  2  I(L^t_{j,1};S^t_j).
	\end{equation}
	Putting (\ref{ineq:36}) and (\ref{ineq:37}) back into (\ref{ineq:32}) completes the proof.
\end{proof}
\subsection{Proof of Theorem \ref{fast-loss}}

\begin{restatetheorem}{\ref{fast-loss}}[Restate]
	Let $n$ and $\tilde{n}$ denote the number of samples available for training the current task and the number of the memory, respectively. Assume that $\ell(\cdot,\cdot) \in [0,1]$, for any $0 < C_2 < \frac{\log 2}{2}$ and $C_1\geq -\frac{\log(2-e^{C_2})}{2C_2}-1$, we have
	\begin{equation*}
		\mathrm{gen}_W \leq C_1\hat{R}+ \sum_{i=1}^{t-1} \sum_{j:U^i_j=1} \frac{ \min \{I(L^i_{j};S^i_j|U^i_j=1), 2 I(L^i_{j,1};S^i_j|U^i_j=1)\} }{\tilde{n}C_2} + \sum_{j=1}^{n}\frac{\min \{I(L^t_{j};S^t_j), 2I(L^t_{j,1};S^t_j) \}}{nC_2}.
	\end{equation*}
	In the interpolating regime that $\hat{R} = 0$, we further have 
	\begin{equation*}
		R \leq  \sum_{i=1}^{t-1} \sum_{j:U^i_j=1} \frac{2 \min \{I(L^i_{j};S^i_j|U^i_j=1), 2 I(L^i_{j,1};S^i_j|U^i_j=1)\} }{\tilde{n}\log 2} + \sum_{j=1}^{n}\frac{2 \min \{I(L^t_{j};S^t_j), 2I(L^t_{j,1};S^t_j) \}}{n\log 2}.
	\end{equation*}
\end{restatetheorem}

\begin{proof}
	Notice that 
	\begin{align}
		R - (1+C_1) \hat{R} = &  \mathbb{E}_{W,\tilde{D}^{1:t-1},S^{1:t-1}, U^{1:t-1}}\Big[\sum_{i=1}^{t-1} \frac{1}{\tilde{n}}\sum_{j:U^i_j=1} \ell(W,\tilde{Z}^i_{j,{1-S^i_j}}) - (1+C_1) \ell(W,\tilde{Z}^i_{j,{S^i_j}})\Big]  \nonumber\\
		& + \mathbb{E}_{W,\tilde{D}^t,S^t} \Big[ \frac{1}{n}\sum_{j=1}^{n}  \ell(W,\tilde{Z}^t_{j,{1-S^t_j}}) - (1+C_1) \ell(W,\tilde{Z}^t_{j,{S^t_j}}) \Big] \bigg\vert \nonumber\\
		= & \mathbb{E}_{U^{1:t-1}} \sum_{i=1}^{t-1} \frac{1}{\tilde{n}}\sum_{j:U^i_j=1} \mathbb{E}_{L^i_j,S^i_j|U^i_j=1} \Big[L^i_{j,\bar{S}^i_j} -  (1+C_1)  L^i_{j,S^i_j}\Big] +  \frac{1}{n}\sum_{j=1}^{n} \mathbb{E}_{L^t_j,S^t_j} \Big[L^t_{j,\bar{S}^t_j} -  (1+C_1)  L^t_{j,S^t_j}\Big] \nonumber\\
		= & \mathbb{E}_{U^{1:t-1}} \sum_{i=1}^{t-1} \frac{1}{2\tilde{n}}\sum_{j:U^i_j=1} \mathbb{E}_{L^i_j,S^i_j|U^i_j=1} \Big[(1+\frac{C_1}{2})(L^i_{j,\bar{S}^i_j}-L^i_{j,S^i_j}) - \frac{C_1}{2}L^i_{j,\bar{S}^i_j} - \frac{C_1}{2}L^i_{j,S^i_j}  \Big] \nonumber\\
		& +  \frac{1}{n}\sum_{j=1}^{n} \mathbb{E}_{L^t_j,S^t_j} \Big[(1+\frac{C_1}{2})(L^t_{j,\bar{S}^t_j}-L^t_{j,S^t_j}) - \frac{C_1}{2}L^t_{j,\bar{S}^t_j} - \frac{C_1}{2}L^t_{j,S^t_j}  \Big] \nonumber\\
		= & \mathbb{E}_{U^{1:t-1}} \sum_{i=1}^{t-1} \frac{1}{2\tilde{n}}\sum_{j:U^i_j=1}  \Bigg( \mathbb{E}_{L^i_j,S^i_j|U^i_j=1} \Big[(-1)^{S^i_j}(2+C_1)L^i_{j,1}-C_1L^i_{j,1}  \Big] \nonumber\\
		& + \mathbb{E}_{L^i_j,S^i_j|U^i_j=1} \Big[-(-1)^{S^i_j}(2+C_1)L^i_{j,0}-C_1L^i_{j,0}  \Big] \Bigg) \nonumber\\
		& + \frac{1}{2n}\sum_{j=1}^{n}  \Bigg( \mathbb{E}_{L^t_j,S^t_j} \Big[(-1)^{S^t_j}(2+C_1)L^t_{j,1}-C_1L^t_{j,1}  \Big]  + \mathbb{E}_{L^t_j,S^t_j} \Big[-(-1)^{S^t_j}(2+C_1)L^t_{j,0}-C_1L^t_{j,0}  \Big] \Bigg) \label{ineq:38}
	\end{align}
	For $i\in[t-1]$, from the proof of Theorem \ref{relia-fast} that $P_{L^i_{j,1}|S^i_j = 0, U^i_j=1} = P_{L^i_{j,0}|S^i_j = 1, U^i_j=1}$ and $P_{L^i_{j,1}|S^i_j = 1, U^i_j=1} = P_{L^i_{j,0}|S^i_j = 0, U^i_j=1}$, we then have $\mathbb{E}_{L^i_{j,1}|S^i_j = 0, U^i_j=1}[L^i_{j,1}] = \mathbb{E}_{L^i_{j,0}|S^i_j = 1, U^i_j=1}[L^i_{j,0}]$ and $\mathbb{E}_{L^i_{j,1}|S^i_j = 1, U^i_j=1}[L^i_{j,1}] = \mathbb{E}_{L^i_{j,0}|S^i_j = 0, U^i_j=1}[L^i_{j,0}]$. Therefore,
	\begin{align}
		\mathbb{E}_{L^i_j,S^i_j|U^i_j=1}\big[(-1)^{S^i_j}\big(2+C_1\big)L^i_{j,1}\big] =&  \frac{\mathbb{E}_{L^i_{j,1}|S^i_j=0,U^i_j=1}\big[\big(2+C_1\big)L^i_{j,1}\big] - \mathbb{E}_{L^i_{j,1}|S^i_j=1,U^i_j=1}\big[\big(2+C_1\big)L^i_{j,1}\big]}{2}  \nonumber\\
		=& \frac{\mathbb{E}_{L^i_{j,0}|S^i_j=1,U^i_j=1}\big[\big(2+C_1\big)L^i_{j,0}\big] - \mathbb{E}_{L^i_{j,0}|S^i_j=0,U^i_j=1}\big[\big(2+C_1\big)L^i_{j,0}\big]}{2}  \nonumber\\
		= &  \mathbb{E}_{L^i_j,S^i_j|U^i_j=1}\big[-(-1)^{S^i_j}\big(2+C_1\big)L^i_{j,0}\big]. \label{ineq:39}
	\end{align}
	For $i=t$, we similarly obtain that 
	\begin{equation}\label{ineq:40}
		\mathbb{E}_{L^t_j,S^t_j} \Big[(-1)^{S^t_j}(2+C_1)L^t_{j,1}-C_1L^t_{j,1}  \Big]  = \mathbb{E}_{L^t_j,S^t_j} \Big[-(-1)^{S^t_j}(2+C_1)L^t_{j,0}-C_1L^t_{j,0}  \Big] 
	\end{equation}
	Plugging inequalities (\ref{ineq:39}) and (\ref{ineq:40}) into (\ref{ineq:38}), we have 
	\begin{align}
		& R - (1+C_1) \hat{R} \nonumber\\
		= & \mathbb{E}_{U^{1:t-1}} \sum_{i=1}^{t-1} \frac{1}{\tilde{n}}\sum_{j:U^i_j=1}  \mathbb{E}_{L^i_j,S^i_j|U^i_j=1} \Big[(-1)^{S^i_j}(2+C_1)L^i_{j,1}-C_1L^i_{j,1}  \Big] + \frac{1}{n}\sum_{j=1}^{n}  \mathbb{E}_{L^t_j,S^t_j} \Big[(-1)^{S^t_j}(2+C_1)L^t_{j,1}-C_1L^t_{j,1}  \Big]. \label{ineq:41}
	\end{align}
	For all $i\in[t-1]$, by applying Lemma \ref{lemmaA.5} with $P=P_{L^i_j,S^i_j|U^i_j=1}$, $Q=P_{L^i_j|U^i_j=1}P_{S^i_j}$, and $f(L^i_j,S^i_j) = (-1)^{S^i_j}\big(2+C_1\big)L^i_{j,1} -  C_1 L^i_{j,1}$, we have 
	\begin{align}
		I^{U^i_j=1}(L^i_j;S^i_j) = & D(P_{L^i_j,S^i_j|U^i_j=1}\Vert P_{L^i_j|U^i_j=1}P_{S^i_j}) \nonumber\\
		\geq & \sup_{C_2>0}\Bigg\{ \mathbb{E}_{L^i_j,S^i_j|U^i_j=1} \Big[(-1)^{S^i_j}C_2\big(2+C_1\big)L^i_{j,1} -  C_2C_1 L^i_{j,1}\Big] \nonumber\\
		&- \log \mathbb{E}_{L^i_j,\tilde{S}^i_j|U^i_j=1}\Big[e^{(-1)^{\tilde{S}^i_j}C_2\big(2+C_1\big)L^i_{j,1} -  C_2C_1 L^i_{j,1}}\Big] \Bigg\} \nonumber\\
		= & \sup_{C_2>0}\Bigg\{ \mathbb{E}_{L^i_j,S^i_j|U^i_j=1} \Big[(-1)^{S^i_j}C_2\big(2+C_1\big)L^i_{j,1} -  C_2C_1 L^i_{j,1}\Big] \nonumber\\
		& - \log \frac{\mathbb{E}_{L^i_j,\tilde{S}^i_j|U^i_j=1}\Big[e^{-2C_2\big(1+C_1\big)L^i_{j,1}} + e^{  2C_2 L^i_{j,1}}\Big]}{2} \Bigg\}. \label{ineq:42}
	\end{align}
	Let $\lambda'_{C_1,C_2} = e^{-2C_2\big(1+C_1\big)L^i_{j,1}} + e^{  2C_2 L^i_{j,1}}$. Similarly, we intend to choose the proper values of $C_1$  and $C_2$ such that the $\log$ term on the RHS can be less than $0$, namely, $\lambda'_{C_1,C_2}\leq 2$. Notice that $\lambda'_{C_1,C_2}$ is convex function of $L^i_{j,1}$, the maximum value of $\lambda'_{C_1,C_2}$ is achieved at the endpoints of $L^i_{j,1}\in[0,1]$. When $L^i_{j,1} = 0$, we obtain $\lambda'_{C_1,C_2} = 2$. When $L^i_{j,1} = 1$, one can select a large enough $C_1\rightarrow \infty$ such that $\lambda'_{C_1,C_2} = e^{-2C_2\big(1+C_1\big)L^i_{j,1}} + e^{  2C_2 L^i_{j,1}}\leq 2$, which yields $C_1\geq - \frac{\log(2-e^{2C_2})}{2C_2}-1$ and $C_2\leq \frac{\log 2}{2}$. Under the above conditions, for any $i\in[t-1]$,
	\begin{equation*}
		\mathbb{E}_{L^i_j,\tilde{S}^i_j|U^i_j=1}\Big[e^{-2C_2\big(1+C_1\big)L^i_{j,1}} + e^{  2C_2 L^i_{j,1}}\Big] \leq 2.
	\end{equation*}
	Putting the above inequality back into (\ref{ineq:42}), we obtain
	\begin{equation}\label{ineq:43}
		\mathbb{E}_{L^i_j,S^i_j|U^i_j=1} \Big[(-1)^{S^i_j}C_2\big(2+C_1\big)L^i_{j,1} -  C_2C_1 L^i_{j,1}\Big] \leq I^{U^i_j=1}(L^i_{j,1};S^i_j).
	\end{equation}
	For $i=t$, by applying the analogous analysis of (\ref{ineq:43}), we can prove that
	\begin{equation}\label{ineq:44}
		\mathbb{E}_{L^t_j,S^t_j} \Big[(-1)^{S^t_j}C_2\big(2+C_1\big)L^t_{j,1} -  C_2C_1 L^t_{j,1}\Big] \leq I(L^t_{j,1};S^t_j).
	\end{equation}
	Substituting inequalities (\ref{ineq:43}) and (\ref{ineq:44}) into (\ref{ineq:41}), we have 
	\begin{equation*}
		\mathrm{gen}_W = R - (1+C_1) \hat{R} + C_1\hat{R} \leq C_1\hat{R}+ \sum_{i=1}^{t-1} \sum_{j:U^i_j=1} \frac{ I(L^i_{j,1};S^i_j|U^i_j=1)}{\tilde{n}C_2} + \sum_{j=1}^{n}\frac{I(L^t_{j,1};S^t_j)}{nC_2}.
	\end{equation*}
	We utilize inequalities (\ref{ineq:36}) and (\ref{ineq:37}) and further get 
	\begin{equation*}
		\mathrm{gen}_W \leq C_1\hat{R}+ \sum_{i=1}^{t-1} \sum_{j:U^i_j=1} \frac{ \min \{I(L^i_{j};S^i_j|U^i_j=1), 2 I(L^i_{j,1};S^i_j|U^i_j=1)\} }{\tilde{n}C_2} + \sum_{j=1}^{n}\frac{\min \{I(L^t_{j};S^t_j), 2I(L^t_{j,1};S^t_j) \}}{nC_2}.
	\end{equation*}
	In the interpolating regime where $\hat{R} = 0$, by setting $C_2\rightarrow \frac{\log 2}{2}$ and $C_1\rightarrow \infty$, we obtain
	\begin{equation*}
		R \leq  \sum_{i=1}^{t-1} \sum_{j:U^i_j=1} \frac{2 \min \{I(L^i_{j};S^i_j|U^i_j=1), 2 I(L^i_{j,1};S^i_j|U^i_j=1)\} }{\tilde{n}\log 2} + \sum_{j=1}^{n}\frac{2 \min \{I(L^t_{j};S^t_j), 2I(L^t_{j,1};S^t_j) \}}{n\log 2}.
	\end{equation*}
	This completes the proof.
\end{proof}

\subsection{Proof of Theorem \ref{fast-var}}
\begin{restatetheorem}{\ref{fast-var}}[Restate]
	Let $n$ and $\tilde{n}$ denote the number of samples available for training the current task and the number of the memory, respectively. Assume that $\ell(\cdot,\cdot) \in \{0,1\}$, for any $0 < C_2 < \frac{\log 2}{2}$ and $C_1\geq -\frac{\log(2-e^{C_2})}{2C_2}-1$, we have
	\begin{equation*}
		\mathrm{gen}_W  \leq C_1 \mathrm{Var}(\gamma)  + \sum_{i=1}^{t-1} \sum_{j:U^i_j=1} \frac{ \min \{I(L^i_{j};S^i_j|U^i_j=1), 2 I(L^i_{j,1};S^i_j|U^i_j=1)\} }{\tilde{n}C_2} + \sum_{j=1}^{n}\frac{\min \{I(L^t_{j};S^t_j), 2I(L^t_{j,1};S^t_j) \}}{nC_2}.
	\end{equation*}
\end{restatetheorem}

\begin{proof}
	By the definition of $\gamma$-variance, we obtain
	\begin{align*}
		\mathrm{Var}(\gamma) = & \mathbb{E}_{W,D^{1:T}} \bigg[ \sum_{i=1}^{t-1}\sum_{j=1}^{\tilde{n}}  \frac{\Big(\ell(W,Z^i_j)-(1+\gamma)\hat{R}^i(W)\Big)^2}{\tilde{n}} + \sum_{j=1}^{n}\frac{\Big(\ell(W,Z^t_j)-(1+\gamma)\tilde{R}^t(W)\Big)^2}{\tilde{n}}   \bigg] \\
		=& \mathbb{E}_{W,D^{1:T}} \bigg[ \sum_{i=1}^{t-1} \frac{1}{\tilde{n}}\sum_{j=1}^{\tilde{n}} \Big( \ell^2(W,Z^i_j)-2(1+\gamma)\ell(W,Z^i_j)\hat{R}^i(W) + (1+\gamma)^2 \big(\hat{R}^i(W)\big)^2 \Big)  \\
		& +   \sum_{j=1}^{n} \frac{1}{n}  \Big(\ell^2(W,Z^t_j)-2(1+\gamma)\ell(W,Z^t_j)\tilde{R}^t(W) + (1+\gamma)^2 \big(\tilde{R}^t(W)\big)^2 \Big) \bigg] \\
		= & \mathbb{E}_{W,D^{1:T}} \bigg[ \sum_{i=1}^{t-1} \frac{1}{\tilde{n}}\sum_{j=1}^{\tilde{n}} \ell(W,Z^i_j) + \sum_{j=1}^{n} \frac{1}{n}\ell(W,Z^t_j)  -2(1+\gamma) \Big(\sum_{i=1}^{t-1} \big(\hat{R}^i(W)\big)^2 + \big(\tilde{R}^t(W)\big)^2 \Big) \\
		&  + (1+\gamma)^2 \Big( \sum_{i=1}^{t-1}  \big(\hat{R}^i(W)\big)^2 + \big(\tilde{R}^t(W)\big)^2 \Big) \bigg] \\
		= & \mathbb{E}_{W,D^{1:T}} \bigg[\sum_{i=1}^{t-1} \frac{1}{\tilde{n}}\sum_{j=1}^{\tilde{n}} \ell(W,Z^i_j) + \sum_{j=1}^{n} \frac{1}{n}\ell(W,Z^t_j)  -(1-\gamma^2) \Big(\sum_{i=1}^{t-1} \big(\hat{R}^i(W)\big)^2 + \big(\tilde{R}^t(W)\big)^2 \Big)  \bigg] \\
		= & \hat{R} -(1-\gamma^2) \mathbb{E}_{W,D^{1:T}} \bigg[\sum_{i=1}^{t-1} \big(\hat{R}^i(W)\big)^2 + \big(\tilde{R}^t(W)\big)^2 \bigg]. 
	\end{align*}
	% \begin{align*}
		%     \mathrm{Var}(\gamma) = & \mathbb{E}_{W,D^{1:T},U,V} \bigg[\frac{1}{kl+n} \sum_{(i,j)\in \{U,V\}\cup\{T,[n]\}} \Big(\ell(W,Z^i_j) - (1+\gamma) L_Z(W)\Big)^2 \bigg] \\
		%     = & \mathbb{E}_{W,D^{1:T},U,V} \bigg[\frac{1}{kl+n} \sum_{(i,j)\in \{U,V\}\cup\{T,[n]\}} \Big(\ell^2(W,Z^i_j) - 2(1+\gamma)\ell(W,Z^i_j) L_Z(W) + (1+\gamma)^2 L^2_Z(W) \Big) \bigg] \\
		%     = &  \mathbb{E}_{W,D^{1:T},U,V} \bigg[\frac{1}{kl+n} \sum_{(i,j)\in \{U,V\}\cup\{T,[n]\}} \ell(W,Z^i_j) \bigg] - 2(1+\gamma) \mathbb{E}_{W,D^{1:T},U,V}\Big[L^2_Z(W)\Big] \\
		%     &  + (1+\gamma)^2  \mathbb{E}_{W,D^{1:T},U,V}\Big[L^2_Z(W)\Big] \\
		%     = &  \hat{R} - (1-\gamma^2) \mathbb{E}_{W,D^{1:T},U,V}\Big[L^2_Z(W)\Big] .
		% \end{align*}
	Since $\ell(\cdot,\cdot)\in\{0,1\}$, we then have $\hat{R}^i(W), \tilde{R}^t(W) \in[0,1]$ for all $i\in[t-1]$, $\big(\hat{R}^i(W)\big)^2 \leq \hat{R}^i(W)$, $\big(\tilde{R}^t(W)\big)^2 \leq \tilde{R}^t(W)$, and 
	\begin{align*}
		\mathrm{gen}_W - C_1\mathrm{Var}(\gamma) = & R - \hat{R} - C_1 \hat{R} + C_1 (1-\gamma^2) \mathbb{E}_{W,D^{1:T}} \bigg[\sum_{i=1}^{t-1} \big(\hat{R}^i(W)\big)^2 + \big(\tilde{R}^t(W)\big)^2 \bigg]\\
		\leq & R - \hat{R} - C_1 \hat{R} + C_1 (1-\gamma^2) \mathbb{E}_{W,D^{1:T}} \bigg[\sum_{i=1}^{t-1}\hat{R}^i(W) + \tilde{R}^t(W) \bigg] \\
		= & R - \hat{R} - C_1 \hat{R} + C_1 (1-\gamma^2) \hat{R} \\
		=  &  R - (1+ C_1 \gamma^2) \hat{R}.
	\end{align*}
	Applying Theorem \ref{fast-loss} with $C_1 =C_1 \gamma^2 $ and $C_2 = C_2$, we have 
	\begin{equation*}
		\mathrm{gen}_W  \leq C_1 \mathrm{Var}(\gamma)  + \sum_{i=1}^{t-1} \sum_{j:U^i_j=1} \frac{ \min \{I(L^i_{j};S^i_j|U^i_j=1), 2 I(L^i_{j,1};S^i_j|U^i_j=1)\} }{\tilde{n}C_2} + \sum_{j=1}^{n}\frac{\min \{I(L^t_{j};S^t_j), 2I(L^t_{j,1};S^t_j) \}}{nC_2},
	\end{equation*}
	under the constraints that $0 < C_2 < \frac{\log 2}{2}$ and $C_1\geq -\frac{\log(2-e^{C_2})}{2C_2\gamma^2}-\frac{1}{\gamma^2}$. This completes the proof.
\end{proof}

\end{document}